\documentclass[lettersize,journal]{IEEEtran}
\usepackage{amsmath,amsfonts}
\usepackage{algorithm}
\usepackage{algpseudocode}

\usepackage{array}

\usepackage{textcomp}
\usepackage{stfloats}
\usepackage{url}
\usepackage{verbatim}
\usepackage{graphicx}
\usepackage{cite}
\usepackage{hyperref}
\hypersetup{hidelinks}
\usepackage{booktabs}
\usepackage[table,xcdraw]{xcolor}
\usepackage{caption}
\usepackage{subcaption}
\usepackage{pifont}
\usepackage{orcidlink}
\usepackage{color}
\usepackage{marvosym}

\newcommand{\toolns}{{\(\text{MA}^{\text{2}}\text{T}\)}}
\newcommand{\tool}{\toolns\space}
\newcommand{\enhanced}{\toolns\space trained\space}

\newcommand{\ie}{\textit{i}.\textit{e}.}
\newcommand{\eg}{\textit{e}.\textit{g}.} 
\newcommand{\Tref}[1]{Tab.~\ref{#1}}
\newcommand{\Eref}[1]{Eq.~(\ref{#1})}
\newcommand{\Fref}[1]{Fig.~\ref{#1}}

\newcommand{\Aref}[1]{Alg.~\ref{#1}}
\newcommand{\etal}{\textit{et al}.}

\begin{document}

\title{Module-wise Adaptive Adversarial Training for\\ End-to-end Autonomous Driving}

\author{
Tianyuan Zhang$^{\orcidlink{0000-0001-9874-6828}}$, 
Lu Wang$^{\orcidlink{0009-0004-1733-4178}}$,  
Jiaqi Kang$^{\orcidlink{0009-0002-2477-5057}}$, 
Xinwei Zhang$^{\orcidlink{0009-0009-2572-3540}}$, 
Siyuan Liang$^{\orcidlink{0000-0002-6154-0233}}$,
Yuwei Chen$^{\orcidlink{0009-0006-2547-5135}}$,~\IEEEmembership{Member,~IEEE,} 
Aishan Liu\Letter$^{\orcidlink{0000-0002-4224-1318}}$,
Xianglong Liu$^{\orcidlink{0000-0001-8425-4195}}$,~\IEEEmembership{Senior Member,~IEEE}

\thanks{T. Zhang, L. Wang, X. Zhang, A. Liu (\Letter corresponding author) and X. Liu are with the School of Computer Science and Engineering, Beihang University, Beijing 100191, China.}

\thanks{J. Kang is with the School of Software, Beihang University, Beijing 100191, China.}

\thanks{S. Liang is with the School of Computing, National University of Singapore, Singapore 117417.}

\thanks{Y. Chen is with the Aviation Industry Development Research Center of China, No.14 Xiao Guan Dong Li, China}

\thanks{(Author T. Zhang and L. Wang contributed equally to this work.)}
}

% The paper headers
\markboth{}%
{Shell \MakeLowercase{\textit{et al.}}: A Sample Article Using IEEEtran.cls for IEEE Journals}
% \IEEEpubid{0000--0000/00\$00.00~\copyright~2021 IEEE}
% Remember, if you use this you must call \IEEEpubidadjcol in the second
% column for its text to clear the IEEEpubid mark.

\maketitle
\begin{abstract}
Recent advances in deep learning have markedly improved autonomous driving (AD) models, particularly end-to-end systems that integrate perception, prediction, and planning stages, achieving state-of-the-art performance. However, these models remain vulnerable to adversarial attacks, where human-imperceptible perturbations can disrupt decision-making processes. While adversarial training is an effective method for enhancing model robustness against such attacks, no prior studies have focused on its application to end-to-end AD models. In this paper, we take the first step in adversarial training for end-to-end AD models and present a novel Module-wise Adaptive Adversarial Training (\toolns). However, extending conventional adversarial training to this context is highly non-trivial, as different stages within the model have distinct objectives and are strongly interconnected. To address these challenges, \tool first introduces Module-wise Noise Injection, which injects noise before the input of different modules, targeting training models with the guidance of overall objectives rather than each independent module loss. Additionally, we introduce Dynamic Weight Accumulation Adaptation, which incorporates accumulated weight changes to adaptively learn and adjust the loss weights of each module based on their contributions (accumulated reduction rates) for better balance and robust training. To demonstrate the efficacy of our defense, we conduct extensive experiments on the widely-used nuScenes dataset across several end-to-end AD models under both white-box and black-box attacks, where our method outperforms other baselines by large margins (+5-10\%). Moreover, we validate the robustness of our defense through closed-loop evaluation in the CARLA simulation environment, showing improved resilience even against natural corruption.
% Our code and demos can be found at the {\href{https://tianyuan2001.github.io/MA2T.github.io/}{\textcolor{blue}{website}}}.

%In this paper, we observe an interconnected relationship between the inputs and outputs of different modules. We thus propose Module-wise Adversarial Spatial Training (\toolns) to adversarially train robust end-to-end AD models. \tool mitigates the impact of adversarial examples by progressively fortifying the model by injecting noise into its interconnected modules. Meanwhile, \tool incorporates dynamic weight accumulation adaptation which balances loss variations across different modules and introduces a temporal factor to accelerate and smooth the loss reduction. We conduct extensive evaluations of \tool on the widely-used nuScenes dataset under both white-box and black-box settings and further assess the closed-loop performance of the reinforced model through CARLA simulation. The experiment results demonstrate that \tool outperforms existing adversarial training methods for end-to-end AD models. Our code and demos can be found at {\href{https://XXXX.github.io}{\textcolor{blue}{Anonymous Website}}}.
\end{abstract}
\section{Introduction}
\IEEEPARstart{R}{ecent} advancements in deep learning have driven significant progress in autonomous driving (AD) models. These models typically involve a series of interconnected tasks, including perception \cite{velasco2020autonomous}, prediction \cite{huang2022survey}, and planning \cite{claussmann2019review}. Traditional approaches often focus on addressing individual tasks in isolation \cite{chen2015deepdriving, cui2019multimodal, song2020pip}, which can lead to issues such as information loss across modules, error accumulation, and feature misalignment \cite{liang2020pnpnet, luo2018fast, sadat2020perceive}. To address these challenges, end-to-end AD models, which unify all components from perception to planning, were proposed to offer a more holistic solution and achieve state-of-the-art performance \cite{hu2023planning, jiang2023vad}.

%more advanced methods integrate multiple tasks within a multi-task learning framework. However, this approach can sometimes result in undesirable ``negative transfer'' \cite{crawshaw2020multi, liu2023bevfusion}. 

% Effective coordination among these modules is crucial for achieving the ultimate decision-making goal. The end-to-end solution has demonstrated state-of-the-art performance \cite{hu2023planning, jiang2023vad}. Therefore, conducting safety research on end-to-end AD models and evaluating and improving their robustness is imperative.

Despite the promising performance, deep-learning-based AD models are highly vulnerable to adversarial attacks, where imperceptible perturbations can significantly degrade the model performance \cite{goodfellow2014explaining, madry2017towards_pgd, liu2019perceptual, liu2020bias, liu2020spatiotemporal, liu2022harnessing, liu2023exploring, guo2023towards}. Previous studies have extensively assessed the adversarial robustness of AD systems in a wide range of sub-tasks (\eg, object detection \cite{eykholt2018robust, huang2020universal,wang2021dual, wang2021adversarial, zhang2023benchmarking} and trajectory prediction\cite{cao2022advdo,zhang2022adversarial, zhang2024lanevil, zhang2024towards}) and even the end-to-end AD models \cite{wu2023adversarial,jiang2024robuste2e, wangattack}. To improve the robustness against adversarial attacks, various defense approaches have been proposed for mitigation \cite{papernot2016distillation, tramer2017ensemble}. Among these, adversarial training \cite{madry2017towards_pgd,liu2021training} has proven particularly effective by incorporating adversarial examples for data augmentation. While there has been considerable research on adversarial training in the context of autonomous driving such as 3D object detection \cite{li2023advmono3d, zhang2024comprehensive}, end-to-end AD models have received relatively little attention. This sparsity of research presents a severe risk to the safety of end-to-end autonomous driving, as it increases their vulnerability to attack.

%Wu \etal  explored adversarial attacks on simple end-to-end models, but their approach primarily treated the model as a classifier with basic outputs (\eg, turn left, turn right), without delving into more advanced architectures. There are two main challenges. First, the distinct objectives of different stages within the model mean that traditional noise injection methods, which focus solely on images, fail to sufficiently train all modules. Second, the robustness of the final output varies across modules, requiring greater attention to those modules that have a more significant impact on the model's overall robustness.

%This highlights the critical need for adversarial safety research on AD models, especially end-to-end solutions, to enhance their robustness.

Therefore, this paper takes the first step in studying adversarial training in the context of end-to-end AD. However, simply extending the existing adversarial training baselines to the context of end-to-end AD is non-trivial owing to the different learning paradigms. In particular, we identify two key challenges impeding robust adversarial training in this scenario: (1) \textit{diverse training objectives}: designing effective adversarial training targets is complex due to the differing objectives of each module in the whole end-to-end pipeline. (2) \textit{different module contributions}: different modules have varying impacts and contributions on the model's final decision robustness. To address these challenges, this paper proposes \textit{Module-wise Adaptive Adversarial Training} (\toolns). As for the issue of diverse training objectives across modules, we design Module-wise Noise Injection, which targets training models with the guidance of overall
objectives rather than each independent module loss. This approach ensures that noise is generated with a holistic view of the model, \ie, using the overall loss for backpropagation instead of focusing on individual module losses that may be contradictory and pose negative impacts on overall decision robustness. To manage the different contributions of modules during training, we introduce Dynamic Weight Accumulation Adaptation,
which adaptively adjusts the loss weights of each module to the overall objectives based on their
contributions during noise injection. In particular, this method incorporates a weight accumulation factor to adjust the descent rates to maintain a balanced training process, which can adaptively control the weight of each module and prevent any module from descending too aggressively during training.

%(accumulated reduction rates) for better balance and robust training. which balances loss reduction across modules during noise injection. This method adjusts descent rates to maintain a balanced training process, preventing any module from descending too quickly or too slowly. Additionally, we incorporate a weight accumulation factor to stabilize weight adjustments, ensuring more consistent and effective training.

% One remaining problem is that learning speed varies for different modules. Drawing inspiration from multi-task learning \cite{guo2018dynamic}, we incorporate a temporal ratio to stabilize weight adjustments, ensuring a more stable and efficient gradient descent.

% Accordingly, we propose \textit{Module-wise Adversarial Spatial Training} (\toolns), which structures the module-wise noise injection process to enhance model robustness and incorporates dynamic weight adaptation to ensure balanced and effective training.
To demonstrate the efficacy of our defense, we conducted extensive experiments in both black-box and white-box attack settings on the widely-adopted nuScenes dataset across several end-to-end AD models, where our \tool achieves superior results compared to common adversarial training methods, with significant improvements (\textbf{+5-10\%}). Moreover, we validate the robustness of our defense through closed-loop evaluation in the CARLA simulation environment, showing improved resilience even against natural corruption. Our \textbf{contributions} can be summarized as follows:

\begin{itemize}
    \item To the best of our knowledge, we are the first to study adversarial training in the context of end-to-end AD.
    
   \item We propose \toolns, which integrates Module-wise Noise Injection and Dynamic Weight Accumulation Adaptation to effectively address the challenges of diverse training objectives and different module contributions.

    \item We conduct extensive experiments to thoroughly evaluate \toolns, demonstrating that it significantly outperforms baseline methods across different adversarial attack methods, achieving absolute improvements of 5-10\%.
\end{itemize}
\section{Preliminaries and backgrounds}

\subsection{Adversarial Attack}

Adversarial attacks introduce slight perturbations to inputs, causing deep learning models to make incorrect predictions. Recent studies have explored various aspects of these attacks \cite{wang2024transferable_liu_1, li2023towards_liu_3, liu2023x_liu_4, zhang2023benchmarking, jiang2023exploring,liu2021training,liu2023towards,zhang2021interpreting,tang2021robustart, zhang2024enhancing}. These attacks are generally categorized into white-box and black-box types. White-box attacks, first introduced by Szegedy \etal \cite{LBFGS}, assume the attacker has full access to the model's internal structure. Goodfellow \etal \cite{FGSM} later developed the Fast Gradient Sign Method (FGSM), while Kurakin \etal \cite{BIM} introduced iterative techniques with the Basic Iterative Method (BIM). Madry \etal \cite{PGD} further refined these methods, leading to the widely recognized Projected Gradient Descent (PGD) attack. In contrast, black-box attacks assume limited access to the model's internal details and utilize methods like Zeroth Order Optimization (ZOO) \cite{chen2017zoo}, which estimates gradients from model outputs, or the Momentum Iterative Fast Gradient Sign Method (MI-FGSM) \cite{MIFGSM}, where adversarial examples generated on a substitute model are used to attack the target model. AutoAttack \cite{croce2020reliable}, a comprehensive attack framework, offers both white-box and black-box modes and is currently considered one of the most effective attacks.

Given an input \(\mathbf{x} \in \mathcal{X}\), where \(\mathcal{X}\) represents the input space (\eg, images), and a corresponding output \(\mathbf{y} \in \mathcal{Y}\), where \(\mathcal{Y}\) denotes the output space (\eg, driving actions), we define a model \(f_\theta: \mathcal{X} \rightarrow \mathcal{Y}\) parameterized by \(\theta\), that maps inputs to outputs. The model's objective is to minimize a loss function \(\mathcal{L}(\mathbf{y}, f_\theta(\mathbf{x}))\) over a dataset \(\mathcal{D} = \{(\mathbf{x}_i, \mathbf{y}_i)\}_{i=1}^{N}\), where \(N\) is the number of samples. This objective can be expressed as:

\begin{equation}
\theta^* = \arg \min_{\theta} \frac{1}{N} \sum_{i=1}^{N} \mathcal{L}(\mathbf{y}_i, f_\theta(\mathbf{x}_i)),
\end{equation}

\noindent where \(\theta^*\) represents the optimal model parameters that minimize the loss over the entire dataset \(\mathcal{D}\).

For a network \(f_\theta\) and an input \(\mathbf{x}\) with ground truth label \(\mathbf{y}\), an adversarial example \(\mathbf{x}^{adv}\) is crafted to cause the network to misclassify, such that:

\begin{equation}
f_\theta(\mathbf{x}^{adv}) \neq \mathbf{y} \quad \text{s.t.} \quad \|\mathbf{x} - \mathbf{x}^{adv}\|_p < \epsilon,
\end{equation}

\noindent where \(\|\cdot\|\) is a distance metric quantifying the perturbation size between \(\mathbf{x}\) and \(\mathbf{x}^{adv}\). Despite this small perturbation, the adversarial example causes the model to predict an incorrect label, specifically \(f_\theta(\mathbf{x}^{adv}) \neq \mathbf{y}\).

\subsection{Adversarial Training}
Adversarial training is currently one of the most effective methods for improving neural network robustness. Initially proposed by Goodfellow \etal \cite{goodfellow2014explaining}, this approach incorporates adversarial examples during training to enhance model resilience. Since then, a variety of adversarial training methods have been developed to further strengthen model defenses. Traditional adversarial training typically blends adversarial examples with standard training data, but this may not provide sufficient protection against a wide range of adversarial attacks. To address this limitation, some researchers have proposed introducing adversarial noise into the intermediate layers of neural networks. For example, Sankaranarayanan \etal \cite{sankaranarayanan2018regularizing} presented a hierarchical adversarial training method that perturbs activations in intermediate layers and computes adversarial gradients based on the previous batch, resulting in stronger regularization. Similarly, Liu \etal \cite{liu2021training} introduced the Adversarial Noise Propagation (ANP) method, which injects diversified noise into hidden layers during training. By generating adversarial noise within the same batch, ANP substantially enhances model robustness. Adversarial training can be expressed as a min-max optimization problem:

\begin{equation}
\label{equ:min-max}
\theta^* = \arg \min_{\theta} \frac{1}{N} \sum_{i=1}^{N} \max_{\|\mathbf{x}_i^{adv} - \mathbf{x}_i\|_p \leq \epsilon} \mathcal{L}(\mathbf{y}_i, f_\theta(\mathbf{x}_i^{adv})),
\end{equation}

\noindent where the inner maximization seeks the worst-case adversarial perturbation \(\mathbf{x}_i^{adv}\) that maximizes the loss, while the outer minimization optimizes the model parameters \(\theta\) to minimize this loss across the training set.

\subsection{End-to-end Autonomous Driving Models}
End-to-end models offer distinct advantages by simplifying the integration of perception and decision-making into a unified framework, thereby minimizing the intricacies of conventional segmented systems. Early end-to-end models seek to consolidate various AD tasks into a single framework to enhance interpretability. A notable example is the P3 series: P3 \cite{sadat2020perceive} innovatively aligns the motion planning costs consistently with perception and prediction estimates, following which, MP3 \cite{casas2021mp3} and ST-P3 \cite{hu2022st} emerge, further advancing end-to-end AD performance by integrating mapping task and learning spatiotemporal features respectively. ST-P3 \cite{hu2022st}, in particular, outperforms previous methods in individual stages. Subsequent models continue to improve learning strategies for AD. For instance, LAV \cite{chen2022learning} not only integrates multi-stage tasks but also incorporates driving experiences from all surrounding vehicles to refine driving strategies. With the advent of Transformer \cite{vaswani2017attention}, the field of AD experienced a notable surge in interest in their application. UniAD \cite{hu2023planning}, the first end-to-end network covering the entire AD stack, employs query-based interactions to facilitate information exchange across tasks and has achieved superior results across the entire stack of tasks. Following this, VAD \cite{jiang2023vad}, also based on Transformer \cite{vaswani2017attention}, uses vectorized scene representations for learning across driving tasks, achieving new state-of-the-art end-to-end planning performance with high efficiency.

End-to-end AD models map sensory inputs \(\mathbf{x} \in \mathcal{X}\) directly to driving actions \(\mathbf{y} \in \mathcal{Y}\). These models typically include \(M\) perception modules \(\{f^{1m}_{\theta}\}_{m=1}^M\), each extracting features \(\mathbf{f}_m \in \mathcal{F}_m\) from \(\mathbf{x}\). These features are then processed by \(K\) prediction modules \(\{f^{2k}_{\theta}\}_{k=1}^K\) to generate predicted states \(\mathbf{p}_k \in \mathcal{P}_k\). Finally, a planning module \(f^3_{\theta}\) combines these predictions to determine the optimal action \(\mathbf{y}\). The model can be expressed as:

\begin{equation}
\label{equ:e2e}
f_\theta(\mathbf{x}) = f^3_{\theta}\left(\{f^{2k}_{\theta}(\{f^{1m}_{\theta}(\mathbf{x})\}_{m=1}^M)\}_{k=1}^K\right),
\end{equation}

\noindent where \(\mathbf{\theta} = (\bigcup_{m=1}^{M} \theta^{1m}) \cup (\bigcup_{k=1}^{K} \theta^{2k}) \cup \{\theta^3\}\) represents the parameters of all modules. The objective is to minimize the loss \(\mathcal{L}(\mathbf{y}, f_\theta(\mathbf{x}))\) across the entire model.

\section{Threat Model}

\subsection{Challenges for Adversarial Training of End-to-end Autonomous Driving models}

Existing adversarial training techniques primarily focus on single-module tasks, where adversarial inputs are generated at the image level. However, applying these methods directly to advanced module-wise end-to-end autonomous driving (AD) models introduces significant challenges. We identify two key challenges that must be addressed to enhance the robustness of these complex models effectively:

\textbf{Challenge \ding{182}: Diverse Training Objectives.} \textit{Designing effective adversarial training targets is complex due to the differing objectives of each module.} In single-module tasks, using the module's specific loss function to train the model adversarially is generally effective. This approach directly targets the module’s weaknesses, making it easier to enhance its robustness. However, in end-to-end AD models, where multiple interconnected modules work together to achieve a final outcome, this strategy falls short. Each module, from perception to planning, has a unique objective and operates under different constraints. Simply focusing on individual module losses during adversarial training may not lead to meaningful improvements in the model's overall robustness. Instead, it is essential to design training targets that consider the collective impact of all modules, ensuring that the adversarial training enhances the resilience of the entire model rather than just isolated components.

\textbf{Challenge \ding{183}: Different Module Contributions.} \textit{Different modules have varying impacts on the model's final robustness.} The ultimate output of an end-to-end AD model, such as the trajectory planning or final driving decision, is heavily influenced by the planning module. However, the robustness of this final output is also dependent on the quality of inputs provided by preceding modules, such as those handling perception and prediction tasks. Each module’s performance can differently affect the overall system. As a result, it is important to identify which modules contribute most significantly to the model’s final robustness and prioritize them during adversarial training. By focusing on fortifying these key modules, the overall stability and safety of the end-to-end AD model can be significantly enhanced, leading to better performance under adversarial conditions.

\subsection{Adversarial Goals}

% In end-to-end AD models, the defender's primary objective is to ensure the model remains robust under the most challenging adversarial conditions, as represented by \Eref{equ:e2e}. The defender seeks to minimize the worst-case loss function \(\mathcal{L}(\mathbf{y}, f_\theta^{adv})\) by:

% \begin{equation}
% \label{equ:min-max-e2e}
% \theta^* = \arg \min_{\theta} \frac{1}{N} \sum_{i=1}^{N} \max_{\|\mathbf{x}_i^{adv} - \mathbf{x}_i\|_p \leq \epsilon, \mathbf{n} \in \mathcal{N}} \mathcal{L}(\mathbf{y}_i, f_\theta(\mathbf{x}_i^{adv}, \mathbf{n})),
% \end{equation}

In end-to-end AD models, the defender's primary objective is to ensure the model remains robust under the most challenging adversarial conditions, as represented by \Eref{equ:e2e}. The defender seeks to minimize the worst-case loss function \(\mathcal{L}(\mathbf{y}, f_{\theta}^{adv})\) by:

\begin{equation}
\label{equ:min-max-e2e}
\theta^* = \arg \min_{\theta} \frac{1}{N} \sum_{i=1}^{N} \max_{\|\mathbf{\delta}_i\|_p \leq \epsilon} \mathcal{L}(\mathbf{y}_i, f_{\theta}(\mathbf{x}_i + \mathbf{\delta}_i)),
\end{equation}

\noindent where \(\mathbf{\delta}_i\) represents the specific adversarial perturbation applied to the \(i\)-th input. 
The perturbation \(\mathbf{\delta}_i = (\bigcup_{m=1}^{M} \delta^{1m}_i) \cup (\bigcup_{k=1}^{K} \delta^{2k}_i) \cup \{\delta^3_i\}\) corresponds to the adversarial noise applied to the perception, prediction, and planning modules for the \(i\)-th input. 
This ensures the model is trained to be robust against adversarial perturbations specific to each input.

% \begin{equation}
% \label{equ:adv}
% f_\theta^{adv}(\mathbf{x}^{adv},\mathbf{n}) = f_{\theta_3}^{adv}\left(\left\{f_{\theta_{2j}}^{adv}\left(\left\{f_{\theta_{1i}}^{adv}(\mathbf{x}^{adv})\right\}_{i=1}^M\right)\right\}_{j=1}^{N}\right),
% \end{equation}

% \subsection{Attack Pipeline}

% The attack pipeline starts with the adversary generating adversarial examples by applying subtle perturbations to the input data, which are designed to cause significant deviations in the model's output while remaining imperceptible. The model then processes these perturbed inputs, propagating through the perception, prediction, and planning modules, ultimately compromising decision-making. The objective is to manipulate the model's behavior in real-world scenarios, such as causing the vehicle to make unsafe decisions—like misidentifying obstacles or taking incorrect turns—especially in critical situations like intersections or pedestrian crossings.

\subsection{Adversary's Capability and Knowledge}
The adversary's capabilities and knowledge vary in different scenarios:

\textbf{White-box Attack.} In a white-box scenario, the adversary has full access to the model's input-output pairs and gradient information, enabling them to craft adversarial examples with detailed insight into the model's behavior. The adversary can also inject noise into various modules of the model, effectively disrupting its architecture.

\textbf{Black-box Attack.} In a black-box scenario, the adversary lacks access to gradient information but can still observe the model's input-output behavior. Although the adversary cannot directly manipulate the model's internals, they possess knowledge of its architecture, allowing them to iteratively generate adversarial examples by applying noise generated from a model with the same architecture. In this scenario, the attacker is limited to adding noise only to the model's most basic input, such as images.

% \subsection{Attack Requirements}

% \subsubsection{Functionality-preserving}
% The attacked model should preserve its original functionality. Given any input without adversarial noise, the model should perform as intended, producing correct outputs consistent with its normal behavior. 

% \subsubsection{Stealthiness}
% The adversarial examples should be indistinguishable from normal inputs to human observers and standard detection mechanisms. 

% \subsubsection{Effectiveness}
% The attack must reliably cause the model to make erroneous decisions, particularly in critical scenarios such as obstacle avoidance or path planning. 
\section{Method}
In this section, we present our proposed approach, \textit{Module-wise Adaptive Adversarial Training} (\toolns), which is composed of two key components: Module-wise Noise Injection and Dynamic Weight Accumulation Adaptation.

\subsection{Module-wise Noise Injection}

To address the challenges posed by diverse training objectives, we designed the Module-wise Noise Injection method. Instead of applying noise at the image level, we inject it directly into each module's input (\eg, adding noise to the image in the perception module) to ensure comprehensive training across all modules. While the typical approach targets each module's loss, this can lead to inconsistent impacts on the model. To mitigate this, we adopt a unified objective to guide training. Our Ablation Study highlights the effectiveness of this method, offering a novel solution to challenge \ding{182}.

% In our \toolns, we adopt the overall model loss as the attack target, offering three key advantages: \ding{182} consistent training across all modules, \ding{183} enhanced robustness of the entire model, and \ding{184} simplified backpropagation, requiring only a single pass rather than multiple passes for each module’s loss.

% Specifically, as shown in \Eref{equ:adv_objective}, we introduce noise \(\delta\) at the input of each module to achieve the effect of module-wise noise injection:

% \begin{equation}
% \label{equ:adv_objective}
% f_\theta(\mathbf{x}, \mathbf{n}) = f_{\theta_3}\left(\{f_{\theta_{2j}}(\{f_{\theta_{1i}}(\mathbf{x}, \delta_{1i})\}_{i=1}^M, \delta_{2j})\}_{j=1}^N, \delta_3\right),
% \end{equation}

% \noindent where \(\mathbf{n} = \{\delta_{1i}\}_{i=1}^M \cup \{\delta_{2j}\}_{j=1}^N \cup \delta_3\) represents the collection of all adversarial noises applied throughout the model.

Specifically, as shown in \Eref{equ:adv_objective}, we introduce noise \(\mathbf{\delta}_i\) at the input of each module to achieve module-wise noise injection:

\begin{equation}
\label{equ:adv_objective}
f^{adv}_\theta(\mathbf{x}_i, \mathbf{\delta}_i) = f_{\theta}^3\left(\{f_{\theta}^{2k}(\{f_{\theta}^{1m}(\mathbf{x}_i, \delta_i^{1m})\}_{m=1}^M, \delta_i^{2k})\}_{k=1}^K, \delta_i^3\right),
\end{equation}

\noindent where \(\mathbf{\delta}_i = (\bigcup_{m=1}^{M} \delta^{1m}_i) \cup (\bigcup_{k=1}^{K} \delta^{2k}_i) \cup \{\delta^3_i\}\) represents the specific adversarial perturbation applied to the \(i\)-th image across the perception, prediction, and planning modules.

During regular training, the total loss \(\mathcal{L}_{total}\) is computed by aggregating the losses from each module. Specifically, it is the sum of the perception module losses \(\mathcal{L}^{1m}\), the prediction module losses \(\mathcal{L}^{2k}\), and the planning module loss \(\mathcal{L}^3\):

\begin{equation}
\label{equ:total-loss}
\mathcal{L}_{total} = \sum_{m=1}^{M} \mathcal{L}^{1m} + \sum_{k=1}^{K} \mathcal{L}^{2k} + \mathcal{L}^3=\sum \mathcal{L}_{j}.
\end{equation}

\noindent where we introduce \(\sum \mathcal{L}_{j}\) to represent the sum of losses across all modules, simplifying the subsequent description. For generating each \(\delta_i\), we use the total loss \(\mathcal{L}_{total}\) as the attack objective. As shown in \Eref{equ:max-loss}, the goal of the adversarial model \(f_\theta^{adv}(\mathbf{x}_{i}, \mathbf{\delta}_i)\) is to maximize the overall loss under adversarial conditions, which is formulated as:

% \begin{equation}
% \label{equ:max-loss}
% \mathbf{n^*} = \arg \max_{\|\mathbf{x}_i^{adv} - \mathbf{x}_i\|_p \leq \epsilon, \mathbf{n} \in \mathcal{N}} \mathcal{L}_{total}(\mathbf{y}_i, f_\theta(\mathbf{x}_i^{adv}, \mathbf{n})),
% \end{equation}

\begin{equation}
\label{equ:max-loss}
\mathbf{\delta}^{*}_{i} = \arg \max_{\|\mathbf{\delta}^i\|_p \leq \epsilon} \mathcal{L}_{total}(\mathbf{y}_i, f_\theta^{adv}(\mathbf{x}_i, \mathbf{\delta}_i)),
\end{equation}

\begin{figure*}[ht]
    \centering
    \includegraphics[width=0.95\linewidth]{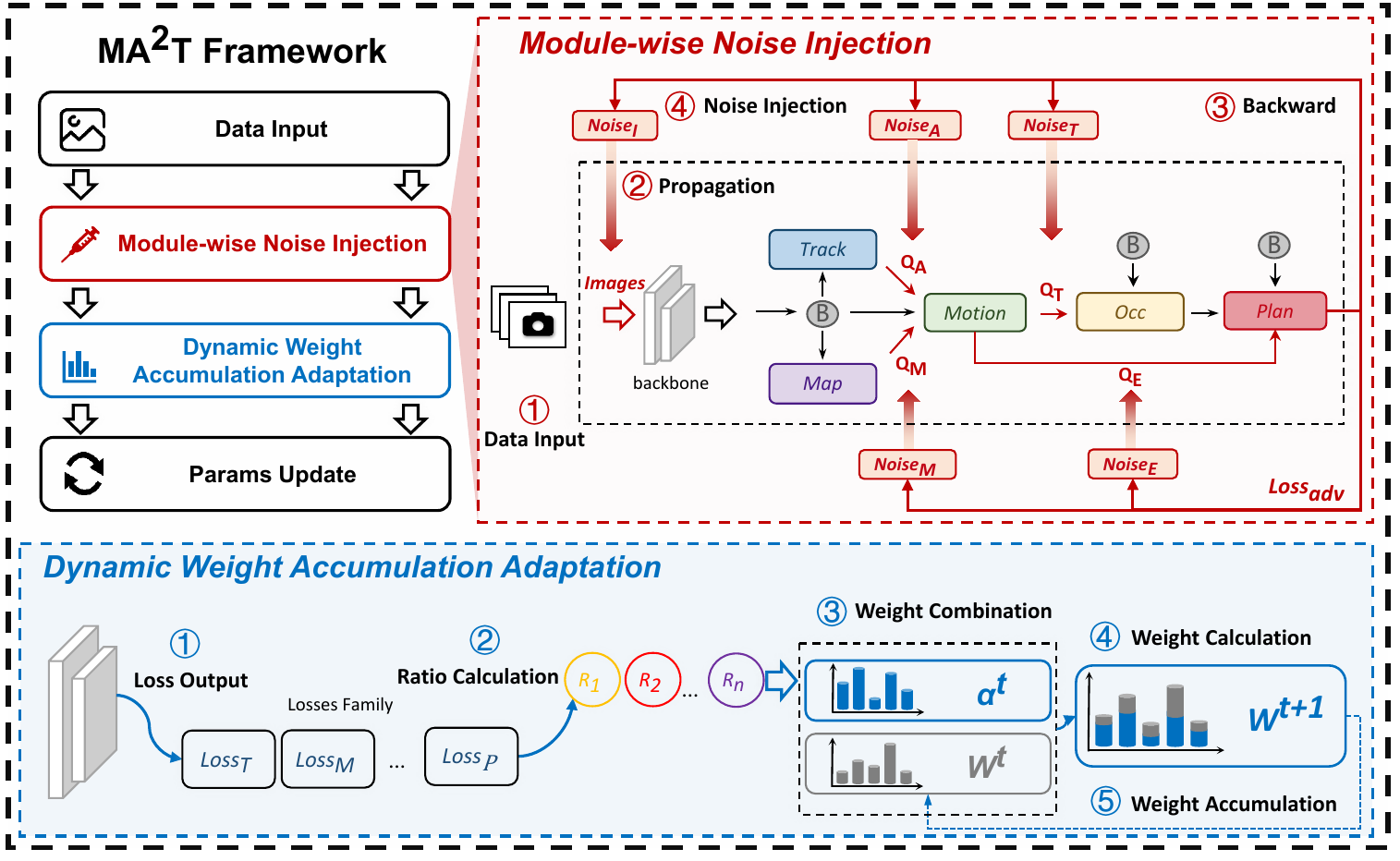}
    \caption{Illustration of \tool (using UniAD as an example). Noise can be introduced at different modules, either directly on the input data or within the connections between modules.}
    \label{fig:framework}
\end{figure*}

This strategy ensures that the targets of attacks when generating noise are $\mathcal{L}_{total}$. The framework of Module-wise Noise Injection is illustrated in \textcolor[RGB]{172, 0, 0}{the red area of \Fref{fig:framework}}, using UniAD \cite{hu2023planning} as an example. The perception modules \(\{f_{\theta}^{1m}\}_{m=1}^M\) include Track and Mapping, while the prediction modules \(\{f_{\theta}^{2k}\}_{k=1}^K\) consist of Motion and Occupancy. The Planning module \(f_{\theta}^3\) then generates the driving actions. Thus, noise can be injected into five distinct modules. The noise injection pipeline proceeds as follows: \ding{182} data input, \ding{183} data propagation, \ding{184} loss backpropagation, and \ding{185} noise injection. Step \ding{186} provides the gradients for noise generation.

\subsection{Dynamic Weight Accumulation Adaptation}

To address the varying module contributions mentioned in challenge \ding{183}, we introduced Dynamic Weight Accumulation Adaptation, which adaptively adjusts the loss weights of each module by incorporating accumulated weight changes. This approach leverages the modules' contribution (accumulated reduction rates) to ensure better balance and more robust training. The loss for each module during forward propagation can be described as \(\mathcal{L}_j^t\), which represents the loss of module \(j\) at time step \(t\).

In multi-task learning, Dynamic Weight Accumulation (DWA) \cite{guo2018dynamic} adjusts the relative size of losses before and after each task to balance the overall loss. Extend the concept of multitasking to multiple modules, the ratio of each module's loss \(\mathcal{L}_j\) at the current time step \(t\) relative to its previous value is calculated as:

\begin{equation}
R_j^t = \frac{\mathcal{L}_j^{t-1}}{\mathcal{L}_j^{t-2}},
\end{equation}

\noindent where ratio \(R_j^t\) captures the relative change in the loss for module \(j\) between two consecutive time steps. To improve this approach, we introduce a scaling factor \(\gamma_j^t\) to represent the significance of the change in the loss ratio for module \(j\) relative to the average change across all modules. This factor is defined as:

\begin{equation}
\alpha_j^t = \frac{N \cdot \gamma_j^t}{\sum_{k=1}^N \gamma_k^t}, \quad \text{where} \quad \gamma_j^t = \exp\left(\frac{R_j^t - \overline{R^t}}{\sigma_{R^t}}\right),
\end{equation}

\noindent where \(\overline{R^t}\) is the mean of the ratios for all modules at time step \(t\), and \(\sigma_{R^t}\) is the standard deviation. Considering relying solely on the last two losses can lead to instability and suboptimal solutions. To mitigate this, we introduced a time decay factor to account for the temporal rates of change in losses. The weight for each module at the next time step \(t+1\) is updated by applying the time decay factor \(r\) to the previous weight and incorporating the newly calculated weight:

\begin{equation}
W_j^{t+1} = r \cdot W_j^t + (1 - r) \cdot \alpha_j^t,
\end{equation}

\noindent where \(W_j^t\) is the weight at the current time step, and \(\alpha_j^t\) is a learning rate adjustment factor for module \(j\). The final total loss for the model at time step \(t+1\) is then calculated by summing the weighted losses from all modules:

\begin{equation}
\mathcal{L}_{total}^{t+1} = \sum_{j=1}^N W_j^{t+1} \cdot \mathcal{L}_j^t.
\end{equation}

This approach ensures the weights adapt dynamically to each module's performance over time, promoting stability and improved overall performance. This exponential adjustment allows the model to prioritize modules with significant performance changes while maintaining overall stability. The pipeline can be outlined as follows: \ding{182} loss output, \ding{183} ratio calculation, \ding{184} weight combination, \ding{185} weight calculation, and \ding{186} weight accumulation. The framework of this approach is illustrated in \textcolor[RGB]{0, 0, 255}{the blue area of \Fref{fig:framework}}.

\subsection{Overall Training}

The overall training process of \tool is outlined in \Aref{alg:tool}. Due to the lengthy training time, we adopt a fine-tuning approach rather than training from scratch. The process begins with a pre-trained model \(f_\theta\), which has already learned standard driving tasks, allowing \tool to focus on adversarial fine-tuning to enhance robustness efficiently.

During fine-tuning, adversarial noise \(\delta_i\) is strategically introduced into various modules using the PGD-\(\ell_\infty\) method, which iteratively optimizes the noise to maximize the loss function within a specified perturbation budget. These noises are injected at different stages—perception, prediction, and planning—across each mini-batch \(B_k\) of the dataset \(\mathcal{D}\). The integration of Module-wise Noise Injection ensures that every module is exposed to adversarial conditions, promoting comprehensive training across the model.

To manage the unequal impact of these noises on different modules, we employ the Dynamic Weight Accumulation Adaptation technique, which dynamically adjusts the noise weights. Unlike traditional methods that update weights at the end of each epoch, our approach updates these weights every 100 batches, allowing the model to quickly adapt and maintain balance across all modules, thereby improving overall robustness.

The training process follows a min-max optimization framework, aiming to minimize the worst-case loss under adversarial conditions, ensuring that the fine-tuned model remains robust against even the most challenging attacks.

\begin{algorithm}[ht]
\caption{Training Procedure of \toolns}
\label{alg:tool}
\begin{algorithmic}[1]
\State \textbf{Input:} Pre-trained model $f_\theta$, dataset $\mathcal{D}$, number of iterations $T$, number of batches per iteration $B$, noise types $\mathbf{\delta}^i = (\bigcup_{m=1}^{M} \delta^i_{1m}) \cup (\bigcup_{k=1}^{K} \delta^i_{2k}) \cup \{\delta^i_3\}$
\State \textbf{Output:} Fine-tuned model parameters $\theta^*$
\State Initialize model parameters $\theta \leftarrow \theta_{\text{pre-trained}}$
\For{each iteration $t = 1, 2, \ldots, T$}
    \For{each batch $b = 1, 2, \ldots, B$}
        \State Generate noise \(\delta^{i}\) for each image
        \State Compute the total loss $\mathcal{L}_{total}^{t} = \sum_{j=1}^N W_j^{t} \cdot \mathcal{L}_j^{t-1}$
        \If{$b \% 100 == 0$}
            \For{each module $j$ in the model}
                \State Compute $R_j^t = \frac{\mathcal{L}_j^{t-1}}{\mathcal{L}_j^{t-2}}$
                \State Compute $\gamma_j^t = \exp\left(\frac{R_j^t - \overline{R^t}}{\sigma_{R^t}}\right)$
                \State Compute $\alpha_j^t = \frac{N \cdot \gamma_j^t}{\sum_{k=1}^N \gamma_k^t}$
                \State Compute $W_j^{t+1} = r \cdot W_j^t + (1 - r) \cdot \alpha_j^t$
            \EndFor
        \EndIf
        \State Backpropagate and update model parameters $\theta$
    \EndFor
\EndFor
\State Return the fine-tuned model parameters $\theta^*$
\end{algorithmic}
\end{algorithm}

In this framework, the Module-wise Noise Injection and Dynamic Weight Accumulation Adaptation are central to enhancing the robustness of end-to-end AD models by ensuring that each module is adversarially trained in a balanced and efficient manner.
\section{Experiments}

We evaluate the effectiveness of our proposed \tool against various adversarial attacks. Following the guidelines from \cite{carlini2019evaluating}, we compare \tool with several commonly-used adversarial defense methods, assessing its performance against both adversarial noise and natural corruption.

\subsection{Experimental Setup}

\textbf{Datasets.} Following \cite{hu2023planning, jiang2023vad}, we conduct our experiments on the nuScenes dataset \cite{caesar2020nuscenes}, a comprehensive AD benchmark that provides richly annotated data across multiple sensor modalities, including cameras from six perspectives, LiDAR, and radar. This dataset is widely used as a standard baseline for evaluating AD tasks, covering a broad spectrum of driving conditions and environments.

\textbf{Models.} We select two representative end-to-end autonomous driving models to validate the effectiveness of \toolns, \ie, UniAD \cite{hu2023planning} and VAD \cite{jiang2023vad}. UniAD is the first model to integrate full-stack autonomous driving tasks, achieving excellent performance on all tasks through plan-oriented collaboration. Unlike the complex construction method of UniAD, VAD learns autonomous driving tasks using vectorized representations, even surpassing UniAD in planning.

\textbf{Metrics.} In our experiments, we employ metrics from \cite{hu2023planning} to ensure consistency and facilitate direct comparison. We evaluate tracking performance using Average Multi-Object Tracking Accuracy (AMOTA $\uparrow$) and assess map alignment through Intersection over Union (IOU $\uparrow$) between predicted and ground-truth maps. Motion forecasting precision is measured by Minimum Average Displacement Error (minADE $\downarrow$), while occupancy accuracy is also evaluated using IOU $\uparrow$. Finally, planning safety and reliability are assessed by Average L2 Error (Avg. L2 Error$ \downarrow$) over the next 3 seconds. 

\textbf{Adversarial Attacks.} For comprehensive robustness evaluation, we consider setting up the model under both white-box and black-box attacks. White-box means that attackers can obtain model information from the victim models, such as gradients, while this information is invisible to black-box attackers. \ding{182} White-box settings. we follow established guidelines \cite{tramer2019adversarial, maini2020adversarial}, incorporating multiple adversarial attacks with various perturbation types and adapting these settings to end-to-end models. We constrain the perturbation under the $\ell_\infty$ norm to 0.2, aligning with common practices on ImageNet. Based on this $\ell_\infty$ setting, we calculate the corresponding $\ell_1$ and $\ell_2$ perturbation constraints—240 and 288,000, respectively—according to the image size of nuScenes. For $\ell_1$ and $\ell_2$ attacks, we use PGD, while for $\ell_\infty$ attacks, we employ PGD, FGSM, and MI-FGSM. The number of iteration steps is set to 5. \ding{183} Black-box settings. We apply the 5 attack methods to generate adversarial examples from the attack models and transfer them to the victim models for testing, selecting the method with the strongest attack effect for reporting.

% Specifically, the \enhanced UniAD and VAD serve as victim models, and their corresponding attack generation models include three categories: the vanilla model, the traditional adversarial trained model, and another autonomous driving model. We choose the method with the strongest attack effect under each black box setting.

\textbf{Adversarial Training Baselines.} We compare four basic adversarial training methods with \toolns. The first method is the origin of adversarial training, \cite{goodfellow2014explaining}, which introduces adversarial losses obtained from FGSM adversarial samples in normal training. \cite{madry2017towards_pgd} proposed a more powerful adversarial training method, namely PGD adversarial training. Based on this, we used $\ell_1$, $\ell_2$ and $\ell_\infty$ norm adversarial training as the other three comparative methods. We use FGSM, $P_{1}$, $P_{2}$ and $P_{\infty}$ to represent the above four methods in the experiments.

\textbf{Implementation Details.} For UniAD, we inject noise at five modules during training, corresponding to the five module interaction paths in the model. The $\ell_\infty$ perturbation constraints for the noise in each module (\ie, Track, Map, Motion, Occ, Plan) are set to 0.8, 0.1, 0.1, 0.1, and 0.1. During the internal maximization, the number of attack iterations is 5, with the training limited to 3 epochs. For VAD, which differs from UniAD in that it uses vectorized representations to infer various driving tasks without clear module divisions, we inject noise into the vectorized representations of detection, mapping, motion, and planning. The corresponding $\ell_\infty$ perturbation constraints for these are all set to 0.1. The iteration count remains at 5, and the training spans 10 epochs. The time decay factor $r$ for each model is set to 0.2. All training and testing are conducted on 8 NVIDIA A800 GPUs (with 80GB of memory each), and the parameters mentioned above are optimized based on multiple training comparisons to achieve the best results.

% \subsubsection{Natural Corruptions}
% To simulate the effects of natural corruptions, we utilize 4 categories of different methods inspired by \cite{hendrycks2019benchmarking}: noise (i.g. gaussian noise, shot noise, and impulse noise), blur (i.g. glass blur, defocus blur, motion blur, and zoom blur), weather (i.g. fog, frost, and snow) and digital distortions (i.g. spatter, contrast, brightness, saturate, jpeg compression, and pixelate). In order to further cover the degree of natural corruptions, 5 severity levels have been set for each type of corruption. We report the average results for four types of corruptions and five severity levels.

\subsection{Main Results}

% Planning Results white
\begin{table}[t]
\centering
\renewcommand\arraystretch{1.2}
\small

\caption{\textbf{Planning results under white-box setting.} The Avg. L2 Error (m) \text{$\downarrow$} with UniAD and VAD models.}
\label{tab:Planning-whitebox}

% UniAD on NuScenes
\begin{tabular}{@{}c@{}}
\label{tab:plan-uniad}
\subfloat[UniAD]{
\resizebox{\linewidth}{!}{
\begin{tabular}{@{}cc|lcccc|c@{}}
\toprule
\textbf{Method}  & Vanilla & F-AT & $P_{1}$ & $P_{2}$ & $P_{\infty}$ & AVG & \toolns(ours)            \\ \midrule
FGSM             & 2.18    & 1.41  & 1.31   & \textbf{1.30}        & 1.47 & 1.37 & \cellcolor[HTML]{EFEFEF} 1.62 \\
MI-FGSM          & 2.47    & 2.06  & 2.44  & 2.28   & 2.13  & 2.23 & \cellcolor[HTML]{EFEFEF} \textbf{1.70} \\
PGD-$\ell_1$      & 1.56    & 1.60  & 1.70  & 1.64   & 1.60        & 1.64 & \cellcolor[HTML]{EFEFEF} \textbf{1.55} \\
PGD-$\ell_2$      & 1.62    & 1.65  & 1.75  & 1.69   & 1.63        & 1.68 & \cellcolor[HTML]{EFEFEF} \textbf{1.52} \\
PGD-$\ell_\infty$ & 2.43    & 1.99  & 2.28  & 2.14   & 2.00  & 2.10 & \cellcolor[HTML]{EFEFEF} \textbf{1.72} \\
AutoAttack  & 2.55 & 2.15 & 2.68 & 2.48 & 2.31 & 2.42 & \cellcolor[HTML]{EFEFEF} \textbf{1.78} \\
% Natural        & 1.10       & 1.67  & 1.20  & 1.67   & 1.96        & 1.76 & \cellcolor[HTML]{EFEFEF} \textbf{1.55} \\
Clean            & \textbf{1.08}   & 1.43  & 1.41  & 1.38   & 1.46        & 1.42 & \cellcolor[HTML]{EFEFEF} 1.28 \\ \bottomrule
\end{tabular}
}}
\end{tabular}

\vspace{0.3cm}

% VAD on NuScenes
\begin{tabular}{@{}c@{}}
\subfloat[VAD]{
\resizebox{\linewidth}{!}{
\begin{tabular}{@{}cc|lcccc|c@{}}
\toprule
\textbf{Method}  & Vanilla & F-AT & $P_{1}$ & $P_{2}$ & $P_{\infty}$ & AVG & \toolns(ours)            \\ \midrule
FGSM             & 0.96    & 1.30  & 1.35  & 1.37   & 1.33        & 1.34 & \cellcolor[HTML]{EFEFEF} \textbf{0.94} \\
MI-FGSM          & 1.12    & 1.20  & 1.34  & 1.33   & 1.31  & 1.30 & \cellcolor[HTML]{EFEFEF} \textbf{1.09} \\
PGD-$\ell_1$      & 1.11    & 1.17  & 1.32  & 1.31   & 1.29  & 1.27 & \cellcolor[HTML]{EFEFEF} \textbf{1.09} \\
PGD-$\ell_2$      & 1.24    & \textbf{1.18}  & 1.32  & 1.32   & 1.30        & 1.28 & \cellcolor[HTML]{EFEFEF} \textbf{1.18} \\
PGD-$\ell_\infty$ & 0.93    & 1.20  & 1.33  & 1.33   & 1.31  & 1.30 & \cellcolor[HTML]{EFEFEF} \textbf{0.89} \\
AutoAttack        & 1.35 & 1.42 & 1.47 & 1.48 & 1.42 & 1.43 & \cellcolor[HTML]{EFEFEF} \textbf{1.24} \\
% Natural        & 1.10       & 1.67  & 1.20  & 1.67   & 1.96        & 1.76 & \cellcolor[HTML]{EFEFEF} \textbf{1.55} \\
Clean            & \textbf{0.73}   & 1.14  & 1.27  & 1.28   & 1.26        & 1.24 & \cellcolor[HTML]{EFEFEF} 1.08 \\ \bottomrule
\end{tabular}}}
\end{tabular}

\end{table}

% Planning Results black
\begin{table}[t]
\centering
\renewcommand\arraystretch{1.3}
% \small

\caption{\textbf{Planning results under black-box setting.} The Avg. L2 Error (m) \text{$\downarrow$} with UniAD and VAD models. ``Att. Gen.'' refers to the attack generation models, while ``Trad. AT'' refers to the Traditional Adversarial Trained models.}
\label{tab:Planning-blackbox}

% UniAD on NuScenes
\begin{tabular}{@{}c@{}}
\subfloat[UniAD]{
\resizebox{\linewidth}{!}{
\begin{tabular}{@{}cc|lcccc|c@{}}
\toprule
\textbf{Att. Gen.}  & Vanilla & F-AT & $P_{1}$ & $P_{2}$ & $P_{\infty}$ & AVG & \toolns(ours)            \\ \midrule
Vanilla             & 1.93    & 1.98  & 2.28  & 2.17   & 2.00        & 2.11 & \cellcolor[HTML]{EFEFEF} \textbf{1.61} \\
% pgdloo 0.2 5
Trad. AT          & 2.23    & 1.92  & 2.17  & 2.15  & 1.88  & 2.03 & \cellcolor[HTML]{EFEFEF} \textbf{1.69} \\
% pgdlooat pgdloo 0.2 5
VAD      & 2.12    & 1.98  & 2.29  & 2.17   & 2.00        & 2.11 & \cellcolor[HTML]{EFEFEF} \textbf{1.61} \\
% vad pgdloo 0.2 5
% PGD-$\ell_2$      & 1.48    & 1.32  & 1.29  & \textbf{1.25}   & 1.27        & 1.28 & \cellcolor[HTML]{EFEFEF} 1.26 \\
% PGD-$\ell_\infty$ & 1.49    & 1.34  & 1.31  & 1.27   & \textbf{1.24}  & 1.29 & \cellcolor[HTML]{EFEFEF} 1.27 \\
% AutoAttack        &        &   &   &    &         &  & \cellcolor[HTML]{EFEFEF} \textbf{} \\
% Natural        & 1.10       & 1.67  & 1.20  & 1.67   & 1.96        & 1.76 & \cellcolor[HTML]{EFEFEF} \textbf{1.55} \\
% Clean            & \textbf{1.03}   & 1.05  & 1.04  & 1.06   & 1.05        & 1.05 & \cellcolor[HTML]{EFEFEF} 1.04 \\ 
\bottomrule
\end{tabular}
}}
\end{tabular}

\vspace{0.3cm}

% VAD on NuScenes
\begin{tabular}{@{}c@{}}
\subfloat[VAD]{
\resizebox{\linewidth}{!}{
\begin{tabular}{@{}cc|lcccc|c@{}}
\toprule
\textbf{Att. Gen.}  & Vanilla & F-AT & $P_{1}$ & $P_{2}$ & $P_{\infty}$ & AVG & \toolns(ours)            \\ \midrule
Vanilla             & 0.79    & 1.15  & 1.28  & 1.29   & 1.26 & 1.25 & \cellcolor[HTML]{EFEFEF} \textbf{0.73} \\
% vad black pgdloo 0.2 5
Trad. AT          & 1.21    & \textbf{1.20}  & 1.31  & 1.32   & 1.31  & 1.29 & \cellcolor[HTML]{EFEFEF} 1.23 \\
% vad pgdlooat pgdloo 0.2 5
UniAD      & 0.76    & 1.14  & 1.28  & 1.28   & 1.26  & 1.24 & \cellcolor[HTML]{EFEFEF} \textbf{0.72} \\
% uniad pgdloo 0.2 5
% PGD-$\ell_2$      & 1.51    & 1.38  & 1.34  & \textbf{1.29}   & 1.31        & 1.33 & \cellcolor[HTML]{EFEFEF} 1.30 \\
% PGD-$\ell_\infty$ & 1.50    & 1.37  & 1.33  & 1.30   & \textbf{1.28}  & 1.32 & \cellcolor[HTML]{EFEFEF} 1.29 \\
% AutoAttack        &        &   &   &    &         &  & \cellcolor[HTML]{EFEFEF} \textbf{} \\
% Natural        & 1.10       & 1.67  & 1.20  & 1.67   & 1.96        & 1.76 & \cellcolor[HTML]{EFEFEF} \textbf{1.55} \\
% Clean            & \textbf{1.04}   & 1.06  & 1.05  & 1.07   & 1.06        & 1.06 & \cellcolor[HTML]{EFEFEF} 1.05 \\ 
\bottomrule
\end{tabular}}}
\end{tabular}

\end{table}

In this section, we evaluate the robustness of the UniAD and VAD models against various perturbation types under both white-box and black-box settings. \textit{We perform five random restarts for attacks for each input to ensure reliability. Our proposed method, \toolns, is trained across five independent runs, and we report the average results from each run.} Given that AD tasks ultimately map raw sensor data to planned trajectory results, we primarily present the results for the Planning task. \textit{ \textit{The results for other tasks are provided in the supplementary materials.}}

% \textbf{White box results:}
\textbf{White-box Results.} The results on nuScenes using UniAD, VAD, \toolns, F-AT, $P_{1}$, $P_{2}$, and $P_{\infty}$ are presented in \Tref{tab:Planning-whitebox}. From these results, we can draw the following observations.

\ding{182} In defending against perturbations (\ie, FGSM, MI-FGSM, PGD-\(\ell_x\)) in white-box scenarios, \tool consistently outperforms other methods, achieving over a 10\% absolute improvement across all five tasks. This underscores the effectiveness of \tool in enhancing the robustness of end-to-end models against a wide range of perturbations.

\ding{183} %The effectiveness of \tool varies across different tasks. 
For the most critical plan, \tool demonstrates significant improvements in resisting various types of attacks. 
%The minADE decreases from 1.3527m to 0.8922 after defense (\Tref{tab:motion}), and 
The Average L2 Error ($\downarrow$) for planning experiences a decrease of 0.64 meters for UniAD and 0.58 meters for VAD (\Tref{tab:plan-uniad}), providing a substantial safety guarantee in real driving scenarios. 
%In upstream tasks, extensive experiments reveal that while UniAD’s \cite{hu2023planning} tracking task shows limited defense, this does not compromise the defense performance of subsequent tasks. For instance, the mapping task's IOU improves by an average of 23\% (\Tref{tab:online}), and occupancy prediction improves by 31\% (\Tref{tab:occupancy}) after defense. For VAD \cite{jiang2023vad}, all tasks benefit from effective defense, with an 8\% increase in detection (\Tref{tab:Multiobject}) and an 11\% increase in mapping (\Tref{tab:online}).

\ding{184} While there is a trade-off between adversarial robustness and standard accuracy \cite{tsipras2018robustness}, leading to slightly lower clean performance compared to the vanilla model, \tool maintains a comparatively high clean performance relative to other adversarial defense strategies.

\textbf{Black-box Results.} We use \enhanced UniAD and VAD as the victim models, with three categories of attack models: the vanilla model with the same architecture, the traditionally adversarial-trained model, and the vanilla model with a different architecture. Based on the results in \Tref{tab:Planning-blackbox}, we can draw the following observations.

\ding{182} Under the black-box settings, the performance degradation of both models is smaller than that of the white-box, but \tool also provides defense against attacks, achieving an average performance improvement of 7.2\% even in the face of unknown attacks, surpassing the 6.0\% of other methods. 

\ding{183} The transfer attack conducted by traditional adversarial training models is the strongest among the three settings, but \tool can also play a defensive role, with UniAD's plan error reduced by 0.2 m and VAD reduced by 0.1 m. 

\ding{184} The attack generated by different model architectures is the weakest, and the corresponding improvement in defense is also relatively small. This still demonstrates \toolns's effectiveness in defending against noise from diverse domains.

% \ding{186} We report the average defense results against natural corruptions, and our \tool is able to resist common disturbances in the natural environment, with an overall performance improvement of 7\%, compared to 0.9\% of other defense methods. This indicates that our method has a more 'intrinsic' nature on enhancing the model.

\subsection{Ablation Study}
\begin{table}[t]
\centering
\renewcommand\arraystretch{0.8}
\Large
\caption{Avg. L2 Error (m) $\downarrow$ of UniAD under different training epochs, PGD-$\ell_\infty$ attack.}
\label{tab:ablation_at_epochs}
\tiny
\resizebox{\linewidth}{!}{
\begin{tabular}{@{}c|clcccc@{}}
\toprule[0.5pt]
\textbf{Epoch} & 0 & 1 & 2 & 3 & 4 & 5 \\ \midrule[0.2pt]
Avg. L2 Error      & 2.43  & 2.11 & 2.07 & 1.72 & 1.69 & 1.68  \\ \bottomrule[0.5pt]
\end{tabular}
}
\end{table}

\begin{table}[t]
\centering
\caption{Avg. L2 Error$\downarrow$ (m) of UniAD under different training perturbation budgets, PGD-$\ell_\infty$ attack.}
\small
\label{tab:ablation_at_eps}

\resizebox{\linewidth}{!}{
\begin{tabular}{@{}c|clccccc@{}}
\toprule[0.7pt]
\textbf{Noise $\epsilon$} & 0 & 0.05 & 0.1 & 0.2 & 0.4  \\ \midrule[0.2pt]
$\text{Noise for Images}^\dagger$      & 2.17  & 2.08 & \textbf{1.72} & 1.84 & 2.03  \\
Noise for Track-Motion      & 2.11  & 1.92 & \textbf{1.72} & 1.89 & 2.04  \\ 
Noise for Map-Motion      & 1.98  & 1.88 & \textbf{1.72} & 1.84 & 1.93  \\ 
Noise for Motion-Occ      & 1.93  & 1.81 & \textbf{1.72} & 1.88 & 1.98  \\ 
Noise for Motion-Plan      & 2.13  & 1.96 & \textbf{1.72} & 1.91 & 2.02  \\ 
\bottomrule[0.7pt]
\end{tabular}
}

\footnotesize{$^\dagger$ The noise value for this row is 8 times the value indicated in the header (\eg, a header value of 0.1 corresponds to an actual noise level of 0.8).}
\end{table}

In this section, we provide some ablation studies to investigate our \tool approach further.

\textbf{Different Noise Budgets and Epochs in \toolns.} This section details our approach to selecting parameter settings in \tool adversarial training, balancing resource constraints with performance outcomes. We use PGD-\(\ell_\infty\) as the evaluation method and report the reduction in planning errors under various training configurations.

First, we address the issue of training epochs. Given the large size of end-to-end models and the time-intensive nature of adversarial training, it's crucial to strike a balance between training time and model performance. For UniAD, one epoch of adversarial training requires an entire day, while the original training process takes 20 epochs. We conducted training for 1 to 5 epochs, with the planning performance results shown in \Tref{tab:ablation_at_epochs}. By the 3rd epoch, the model already demonstrates strong defense capabilities. As training continues, additional epochs result in diminishing returns, with minimal performance improvement despite significant time investment.

Regarding the perturbation budget, we adjusted and compared the noise at each stage by controlling variables, keeping other perturbation settings fixed. Using UniAD as an example, we evaluated the impact of noise at five different stages on model robustness under varying perturbation constraints during training, with planning results as the evaluation metric. Specifically, the baseline \(\ell_\infty\) perturbation constraints for the noise in each module (\ie, Track, Map, Motion, Occ, Plan) were set to 0.8, 0.1, 0.1, 0.1, and 0.1, respectively. We then adjusted the size of only one noise at a time and observed the resulting changes in the trained model. As shown in \Tref{tab:ablation_at_eps}, the contribution of perturbation constraints to model robustness generally follows a convex pattern—both excessively small or large constraints lead to decreased performance. Based on these observations, we selected the optimal adversarial training settings for both models.

% \vspace{0.3cm}

% \subfloat[Noise for Track-Motion]{
% \tiny
% \resizebox{\linewidth}{!}{
% \begin{tabular}{@{}c|clccccc@{}}
% \toprule[0.4pt]
% \textbf{$\epsilon$} & 0.125 & 0.25 & 0.5 & 1 & 2 & 3 & 4 \\ \midrule[0.2pt]
% PGD-$\ell_2$      & 1.30 & 1.37 & 1.49 & 1.65 & 1.85 & 2.07 & 2.32 \\ \bottomrule[0.4pt]
% \end{tabular}
% }
% }

% \vspace{0.3cm}

% \renewcommand\arraystretch{1.5}
% \subfloat[Noise for Map-Motion]{
% \resizebox{\linewidth}{!}{
% \begin{tabular}{@{}c|clccccc@{}}
% \toprule[0.6pt]
% \textbf{$\epsilon$} & 2/255 & 4/255 & 8/255 & 16/255 & 24/255 & 32/255 & 40/255 \\ \midrule[0.3pt]
% PGD-$\ell_\infty$   & 1.42  & 1.50  & 1.62  & 1.78   & 1.95   & 2.15   & 2.37   \\ \bottomrule[0.6pt]
% \end{tabular}
% }
% }

% \subfloat[Noise for Motion-Occupancy]{
% \resizebox{\linewidth}{!}{
% \begin{tabular}{@{}c|clccccc@{}}
% \toprule[0.6pt]
% \textbf{$\epsilon$} & 2/255 & 4/255 & 8/255 & 16/255 & 24/255 & 32/255 & 40/255 \\ \midrule[0.3pt]
% PGD-$\ell_\infty$   & 1.42  & 1.50  & 1.62  & 1.78   & 1.95   & 2.15   & 2.37   \\ \bottomrule[0.6pt]
% \end{tabular}
% }
% }

% \subfloat[Noise for Motion-Plan]{
% \resizebox{\linewidth}{!}{
% \begin{tabular}{@{}c|clccccc@{}}
% \toprule[0.6pt]
% \textbf{$\epsilon$} & 2/255 & 4/255 & 8/255 & 16/255 & 24/255 & 32/255 & 40/255 \\ \midrule[0.3pt]
% PGD-$\ell_\infty$   & 1.42  & 1.50  & 1.62  & 1.78   & 1.95   & 2.15   & 2.37   \\ \bottomrule[0.6pt]
% \end{tabular}
% }
% }

\textbf{Effectiveness of the Module-wise Noise Injection.}
As a core component of our \toolns, the Module-wise Noise Injection is important in bolstering adversarial defense. To better understand its impact, we conduct a series of experiments to evaluate its effectiveness.

We specifically examine the effects of selectively injecting or omitting noise to various modules to assess the robustness of the final model. These experiments are performed using the UniAD model on nuScenes dataset, employing a white-box PGD-$\ell_\infty$ attack with a perturbation budget of $\epsilon=0.2$, 5 steps, and a step size of $\alpha=\epsilon/5$. As depicted in \Fref{fig:framework}, we experiment with five different types of noise. The default setting included injecting all five types, and we explore settings where only one type of noise is either removed or retained during adversarial training to gauge its effect.

\begin{figure}[t]
\centering
% \vspace{-0.07in}
	\begin{subfigure}{0.45\linewidth}
		\centering
        \includegraphics[width=1.0\linewidth]{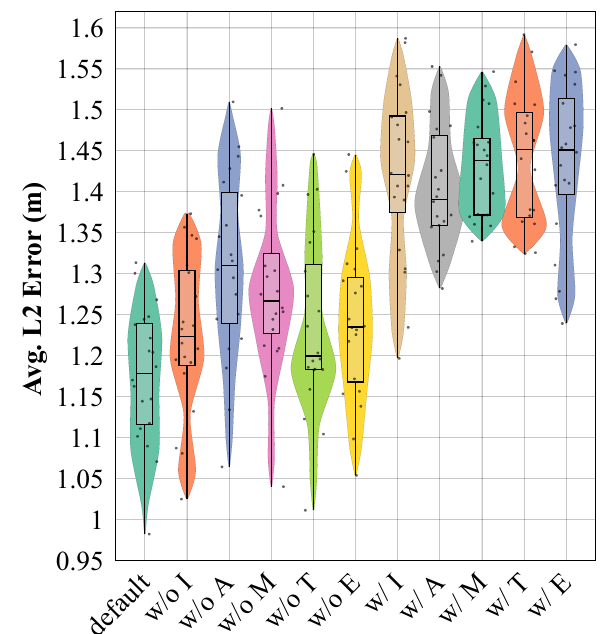}
		\caption{}
		\label{subfig:ablation-a}%文中引用该图片代号
	\end{subfigure}
	\begin{subfigure}{0.48\linewidth}
		\centering
		\includegraphics[width=1.0\linewidth]{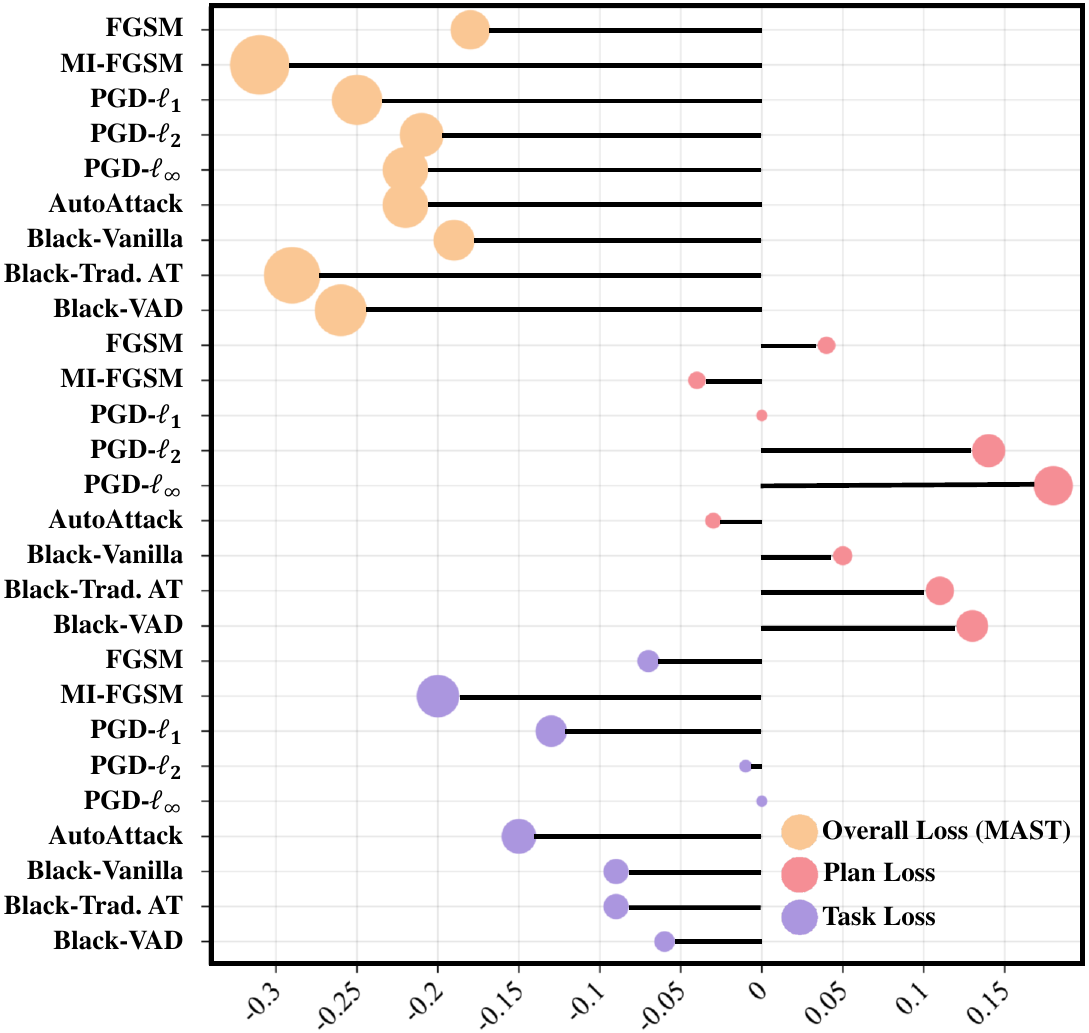}
		\caption{}
		\label{subfig:ablation-b}%文中引用该图片代号
	\end{subfigure}

\caption{Defense results under adaptive white-box attacks.}
\label{fig:d3}
\end{figure}

Given the variability in results across different training sessions, we conducted 20 repeated experiments for each setting and visualized the Avg. L2 Error with a violin plot, as shown in \Fref{subfig:ablation-a}. Overall, the results indicate that the most effective defense is achieved when all types of noise are injected. Additionally, as the number of modules injected with noise increases, the defense effect significantly improves. Specifically, we find that when only a single noise is injected, the noise from the Track-Motion branch contributes the most to the robustness of the model, while when only a single noise is discarded, the planning error is maximized after discarding the noise from the Track-Motion branch. This implies the key and fragile characteristics of the Track-Motion interface.

% \begin{figure}[!]
%     \centering
%     \includegraphics[width=0.95\linewidth]{Images/Ablation_study_noiose_injection.pdf}
%     \caption{Avg. L2 Error (m) $\downarrow$ of UniAD under white-box PGD-$\ell_\infty$. The horizontal axis represents different noise injection methods: ``w/'' indicates injecting only a specific type of noise (e.g., ``w/ I'' means injecting only \textit{$\text{Noise}_\text{I}$}), while ``w/o'' indicates omitting a specific type of noise (e.g., ``w/o I'' means not injecting \textit{$\text{Noise}_\text{I}$} but injecting all other types of noise).}

%     \label{fig:ablation_noise_injection}
% \end{figure}

\textbf{Effectiveness of Dynamic Weight Accumulation Adaptation.} To assess the effectiveness of Dynamic Weight Accumulation Adaptation (DWAA), we conducted experiments by removing this component. Using the UniAD model on the nuScenes dataset, we applied a white-box PGD-$\ell_\infty$ attack with a perturbation budget of $\epsilon=0.2$, 5 steps, and a step size of $\alpha=\epsilon/5$. For training, we set $\ell_\infty$ perturbation constraints for each module (\ie, Track, Map, Motion, Occ, Plan) at 0.8, 0.1, 0.1, 0.1, and 0.1, respectively. The attack steps were set to 5, with fine-tuning over 3 epochs.

\begin{figure}
    \centering
    \includegraphics[width=0.95\linewidth]{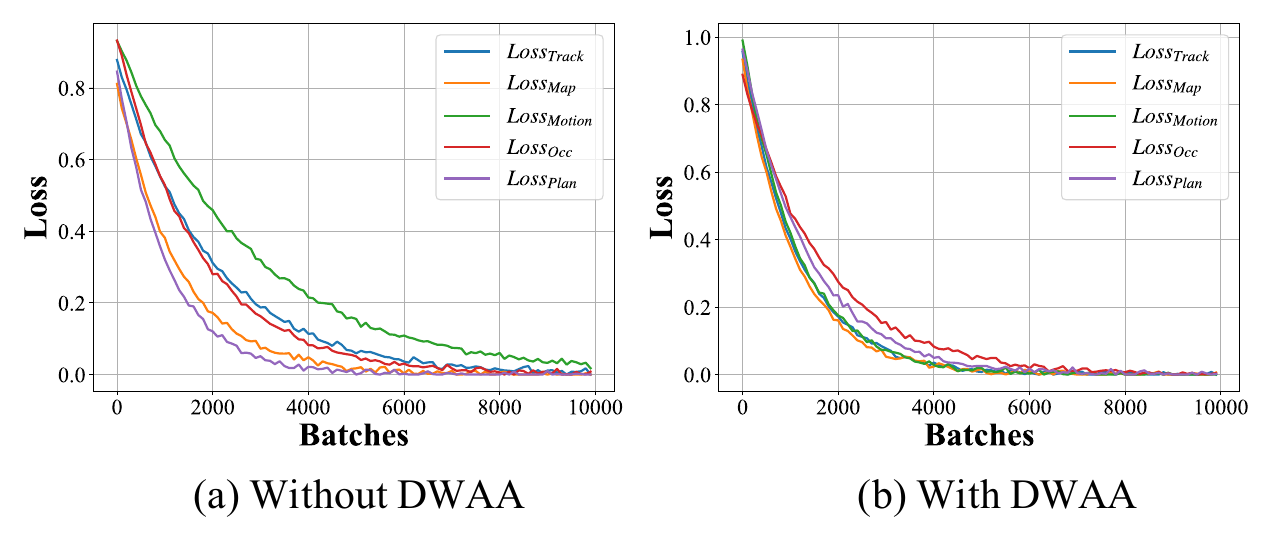}
    \caption{The detailed loss trend with/without Dynamic Weight Accumulation Adaptation. (All losses are proportionally scaled down to the 0-1 range.)}
    \label{fig:loss_decline_compare}
\end{figure}

The results show that the Avg. L2 Error increased from 1.23 to 1.31, indicating a more significant negative impact. Furthermore, as shown in \Fref{fig:loss_decline_compare}, with DWAA, losses nearly converge by 6,000 batches, whereas without DWAA, losses continue to decline until 8,000 batches, and the plan module approaches convergence as early as 4,000 batches. These findings confirm the effectiveness of DWAA in accelerating loss reduction, balancing loss allocation across modules, and enhancing overall model performance.

\textbf{Construction of Noise Objective Function in \toolns.} In \toolns we adopt the overall loss of each task as the objective function for internal maximization. Here we conduct ablation analysis on various objective function constructions and demonstrate the superiority of our design. Since we inject noise into modules during adversarial training, a natural idea is to use each task loss as the optimization target of the corresponding noise. In addition, we also considered using the final plan loss as the target for the internal maximization. We compared the defense results of these two adversarial training methods and our \tool against all attacks in the main experiments. \Fref{subfig:ablation-b} shows the decrease of plan L2 error (m) after applying \tool defense, where it's evident that \tool achieves the greatest error reduction. This demonstrates that thorough training of \tool across all tasks is crucial for the enhancement of end-to-end autonomous driving models.

% \subsection{Adaptive white-box attacks for \toolns}

% TODO

% MW ATTACK

% TASK ORIENTED ATTACK .. 

\subsection{Adaptive White-box Attacks for \toolns}
\begin{figure}[t]
\centering
% \vspace{-0.07in}
	\begin{subfigure}{0.49\linewidth}
		\centering
        \includegraphics[width=1.0\linewidth]{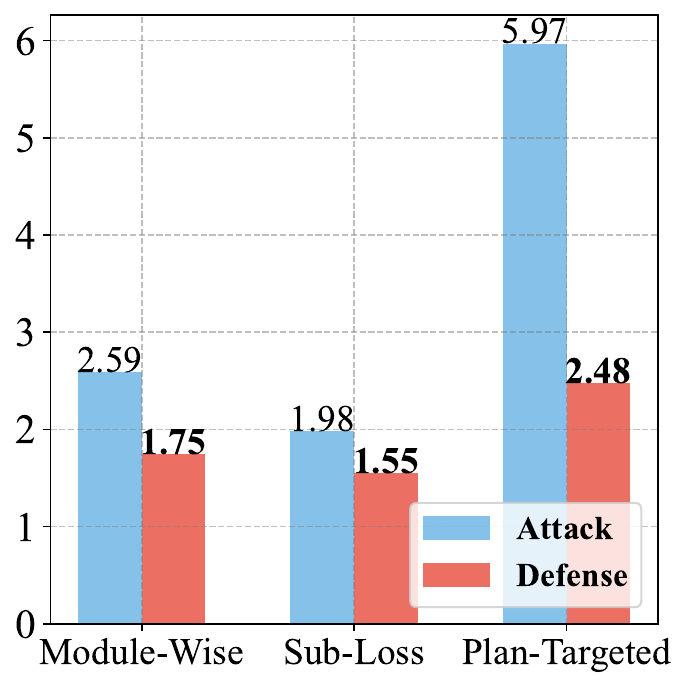}
		\caption{UniAD}
		\label{subfig:d3-unadi}%文中引用该图片代号
	\end{subfigure}
	\begin{subfigure}{0.49\linewidth}
		\centering
		\includegraphics[width=1.0\linewidth]{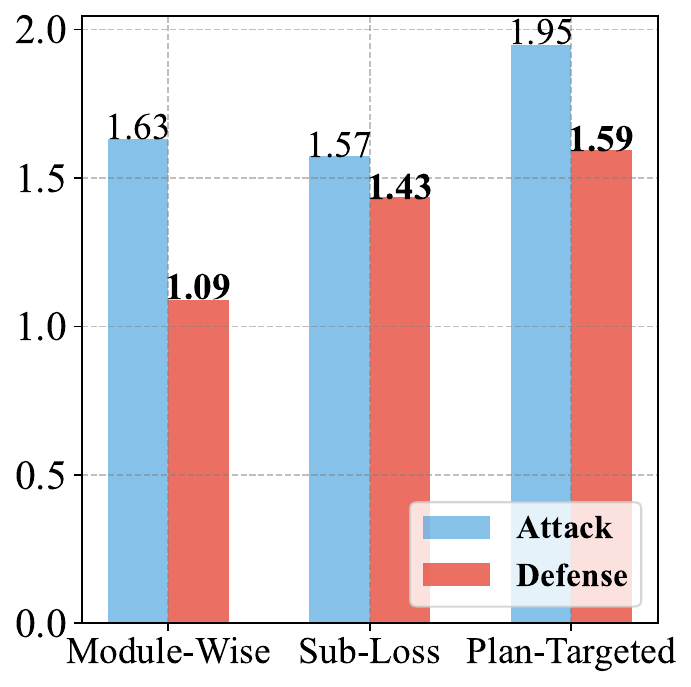}
		\caption{VAD}
		\label{subfig:d3-vad}%文中引用该图片代号
	\end{subfigure}

\caption{Defense results of plan's Avg. L2 Error (m) $\downarrow$ under three adaptive white-box attacks.}
\label{fig:adaptive_attack}
\end{figure}

\textbf{Adaptive Attack Design.} In addition to the commonly used adversarial attacks, we also evaluate the performance of our \tool against white-box attacks specifically designed for it, intending to provide a more thorough analysis. Starting from \toolns, we also design specific attack methods based on modular noise injection. We customize three different types of attacks by utilizing the rich task characteristics of end-to-end autonomous driving models. Our attack is consistent with the PGD-$\ell_\infty$ attack in the main experiments, which uses the $\ell_\infty$ norm constraint, with a perturbation constraint of 0.2 for each module and 5 iterations.

\textbf{\ding{182} Module-wise Attack.} This attack method mirrors the internal maximization process used in our training approach, extracted as a distinct attack strategy. During model inference, noise is injected into each module, with the model's own training target serving as the attack objective. 

\textbf{\ding{183} Sub-loss Attack.} In this approach, targeted attacks are conducted at different stages of the end-to-end model. Noise is injected module by module, with each module’s noise specifically targeting its corresponding task. The sub-loss of the respective task is used as the optimization objective for the corresponding noise. Specifically, the objective function of each noise is the loss of the nearest neighbor task.

\textbf{\ding{184} Plan-targeted Attack.} This attack focuses on the critical planning task in AD. Noise is injected into each subtask during model inference, but all noise is directed toward the final planning output, with the planning loss serving as the goal for all noise optimization.

\textbf{Results and Analysis.}
We evaluated \toolns's defense effectiveness to three adaptive attack methods on UniAD and VAD models. The most critical plan Avg. L2 Error (m) $\downarrow$ is shown in \Fref{fig:adaptive_attack}. 

\ding{182} Comparison of results after attack and defense clearly demonstrates that \tool effectively resists these carefully designed attacks. This further confirms that \tool is not limited to specific attack targets but provides comprehensive defense enhancement across the entire model pipeline. 

\ding{183} \tool shows varying effectiveness against different types of attacks. \tool shows relatively low defense performance improvement against Sub-Loss Attack in two models, but achieves significant error reduction against the other two types of attacks, where UniAD obtains an error reduction of 3.49m (58\%) against Plan-Targeted Attack, and VAD 0.54m against Module-Wise Attack.

\ding{184} The attack intensity of Plan-Targeted Attack is the most severe among the three types of attacks, which is consistent with intuition. In addition, the Module-Wise Attack far outperforms the Sub-Loss attack, indicating that the latter's attack strategy is decentralized and fails to effectively interfere with key parts of the model, which plays a guiding role in the design of \toolns.

\ding{185} While both models are 
extremely susceptible to the adaptive attacks, VAD exhibits a more resilient architecture, possibly due to its vectorized design.

%This enhancement method alsof conforms to the design concept of end-to-end AD, which is the collaboration of various tasks to promote reliable decision-making.

\section{Closed-loop Simulation Evaluation}
\begin{figure}
    \centering
    \includegraphics[width=0.99\linewidth]{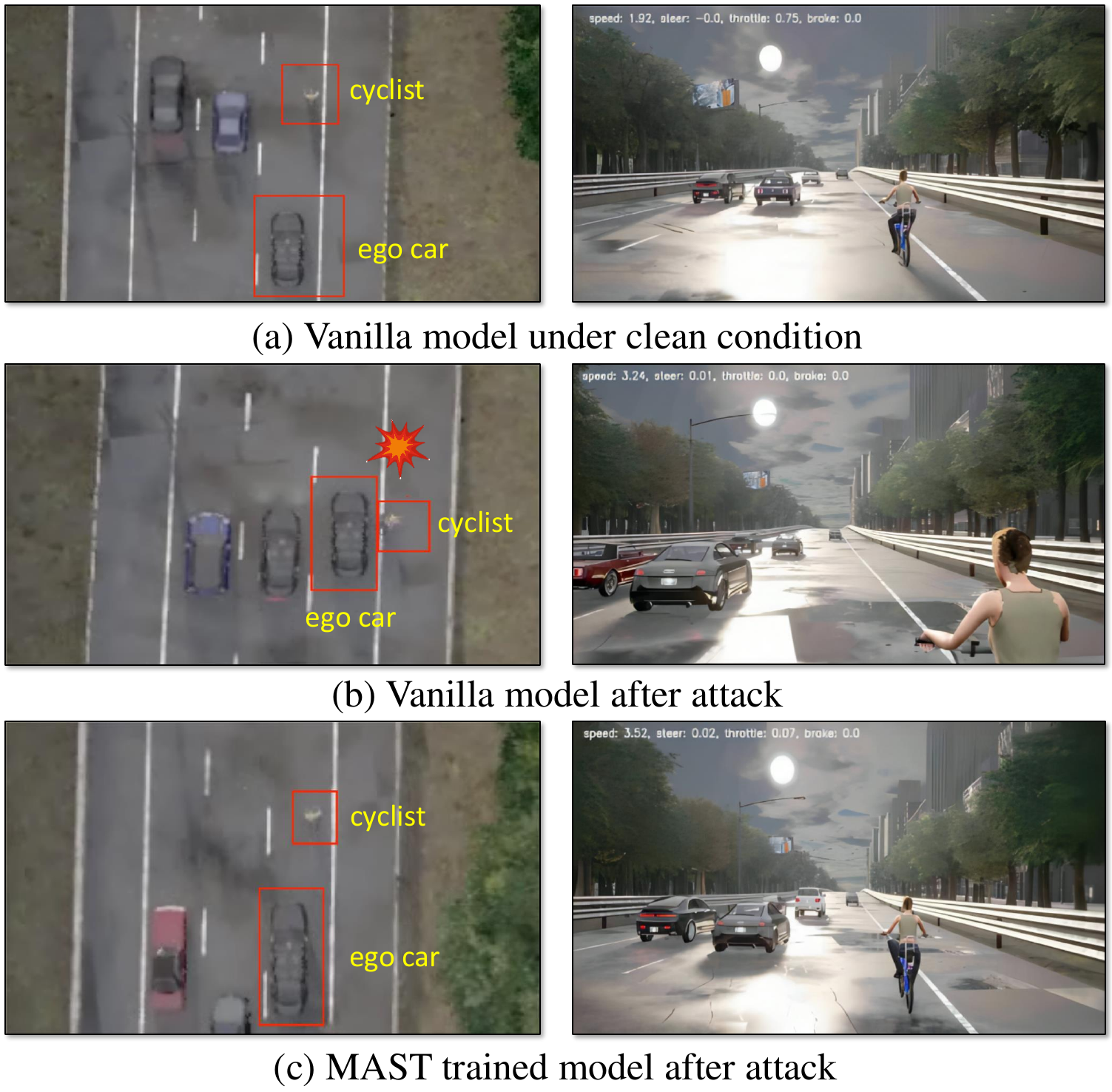}
    \caption{Comparison images of the (a) vanilla model under clean conditions, (b) after attack, and (c) \enhanced model after attack. The images show that the original model slows down and avoids when there is a bicycle ahead. However, after an attack, the ego vehicle directly collides with the bicycle. When using \enhanced model during the attack, the ego vehicle adjusts to the left to avoid the bicycle.}
    \label{fig:bench2drive_compare}
\end{figure}

In this section, we conduct a closed-loop evaluation using the CARLA simulator \cite{dosovitskiy2017carla}, integrating the state-of-the-art UniAD and VAD models. Following the methodology outlined in \cite{jia2024bench2drive}, we input images directly into the models as perception data to assess their decision-making performance in realistic driving scenarios. To validate the real-world effectiveness of our \toolns, we performed black-box attack experiments.

\subsection{Experimental Setup}

\textbf{Simulation Environment Setup.} We set up the simulation environment using CARLA 0.9.15 and Leaderboard v2, focusing on the long-distance routes in Town12 and Town13. Additionally, to ensure comprehensive testing, we also evaluated on other shorter routes from CARLA's official dataset.

\textbf{Model Integration.} Since the UniAD and VAD models do not natively support CARLA simulation, we leveraged the data formats and APIs from Bench2Drive to integrate their perception and decision-making modules with CARLA’s vehicle control system, enabling seamless closed-loop control.

\textbf{Black-box Attack Implementation.} After integrating the models, we conducted black-box attacks on the UniAD and VAD models. Specifically, we used data collected from CARLA to train on different images, generating a universal noise that is applicable to various images. During the testing phase, while running the CARLA simulator, we applied the generated noise to the input images to execute the black-box attacks on the models.

\textbf{Metrics.} The evaluation metrics included the Driving Score $\uparrow$, and task-specific success rates such as Merging (\%) $\uparrow$, Overtaking (\%) $\uparrow$, Emergency Brake (\%) $\uparrow$, and Traffic Sign (\%) $\uparrow$ recognition. These metrics provide a comprehensive view of the models' decision-making capabilities and robustness across various driving scenarios.

\textbf{Implementation Details.} We conducted \tool adversarial training on the UniAD and VAD models separately to enhance robustness. The $\ell_\infty$ perturbation constraints for the noise in each module (\ie, Track, Map, Motion, Occ, Plan) are set to 0.8, 0.1, 0.1, 0.1, and 0.1. The number of attack iterations is 5, with the training limited to 3 epochs.

\subsection{Closed-loop Simulation Results}

To validate the robustness of the adversarially trained models, we tested the \enhanced models on CARLA by applying the generated noise to the input images. The performance comparison between the (a) vanilla models under clean conditions, (b) vanilla models after the attack, and (c) \enhanced models after the attack are illustrated in \Fref{fig:bench2drive_compare}.

\ding{182} Under black-box attacks, the performance of the UniAD and VAD models significantly declined. For example, the driving score of the VAD dropped from 37.72\% to 25.64\%, and the driving score of the UniAD decreased from 39.42\% to 37.91\%. The task-specific performance also suffered. For instance, the merging success rate of the UniAD fell from 4.11\% to 1.25\%.

\ding{183} \tool markedly improved the models' robustness against these attacks. After training, the driving score of the UniAD increased to 38.86\%, and the driving score of the VAD improved to 26.87\%. These results demonstrate that the \tool method is highly effective in mitigating the impact of adversarial attacks, nearly restoring the models' original performance levels. For detailed performance metrics, please refer to \Tref{tab:closed-loop}.

\begin{table}[t]

\centering
\renewcommand\arraystretch{1.2}
\caption{Closed-loop simulation evaluation results.}
\Large
\resizebox{\linewidth}{!}{
\begin{tabular}{@{}cc|ccccc|c@{}}
\toprule[0.8pt] 
\textbf{Method} & Driv. $\uparrow$ & Merg. (\%) $\uparrow$ & Over. (\%) $\uparrow$ & Emer. (\%) $\uparrow$ & Traf. (\%) $\uparrow$ \\ \midrule
$\text{UniAD}^{*}$        & 39.42 & 4.11  & 12.50  & 14.54 & 18.54 \\
$\text{UniAD}^{\dagger}$ & 37.91 & 1.25  & 6.67  & 7.27 & 17.64\\
$\text{UniAD}^{\ddagger}$ & 38.86 & 3.33  & 10.32  & 27.41 & 27.41 \\
$\text{VAD}^{*}$ & 37.72 & 12.50 & 17.50  & 14.54 & 25.55 \\
$\text{VAD}^{\dagger}$ & 25.64 & 5.56 & 0.00  & 0.00 & 20.00 \\
$\text{VAD}^{\ddagger}$ & 35.39 & 12.5 & 16.7  & 11.8 & 22.68 \\
\bottomrule[0.8pt]
\end{tabular}
}
\raggedright
\footnotesize{$^*$ Performance of vanilla models under clean conditions. \\$^\dagger$ Performance of vanilla models after the attack. \\$^\ddagger$ Performance of enhanced models after the attack.}\\

\label{tab:closed-loop}
\end{table}
\section{Discussion and Analysis}

\subsection{Module Contribution Analysis}
\begin{figure*}[!]
\label{Fig:zhexian}
\centering
    \begin{subfigure}[b]{0.32\textwidth}
        \centering
        \includegraphics[width=\textwidth]{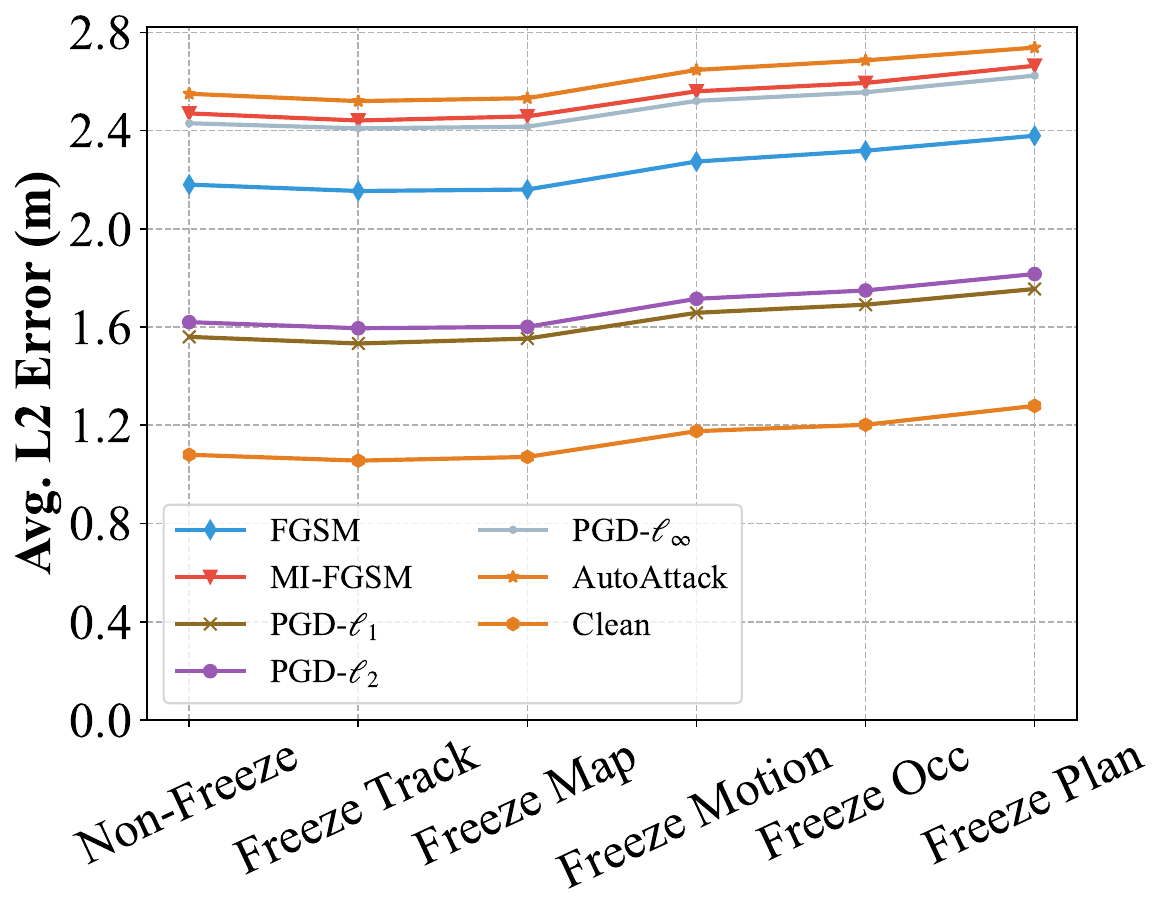}
        \caption{Module Contribution}
        \label{subfig:module-freeze}
    \end{subfigure}
    \hfill
    \begin{subfigure}[b]{0.32\textwidth}
        \centering
        \includegraphics[width=\textwidth]{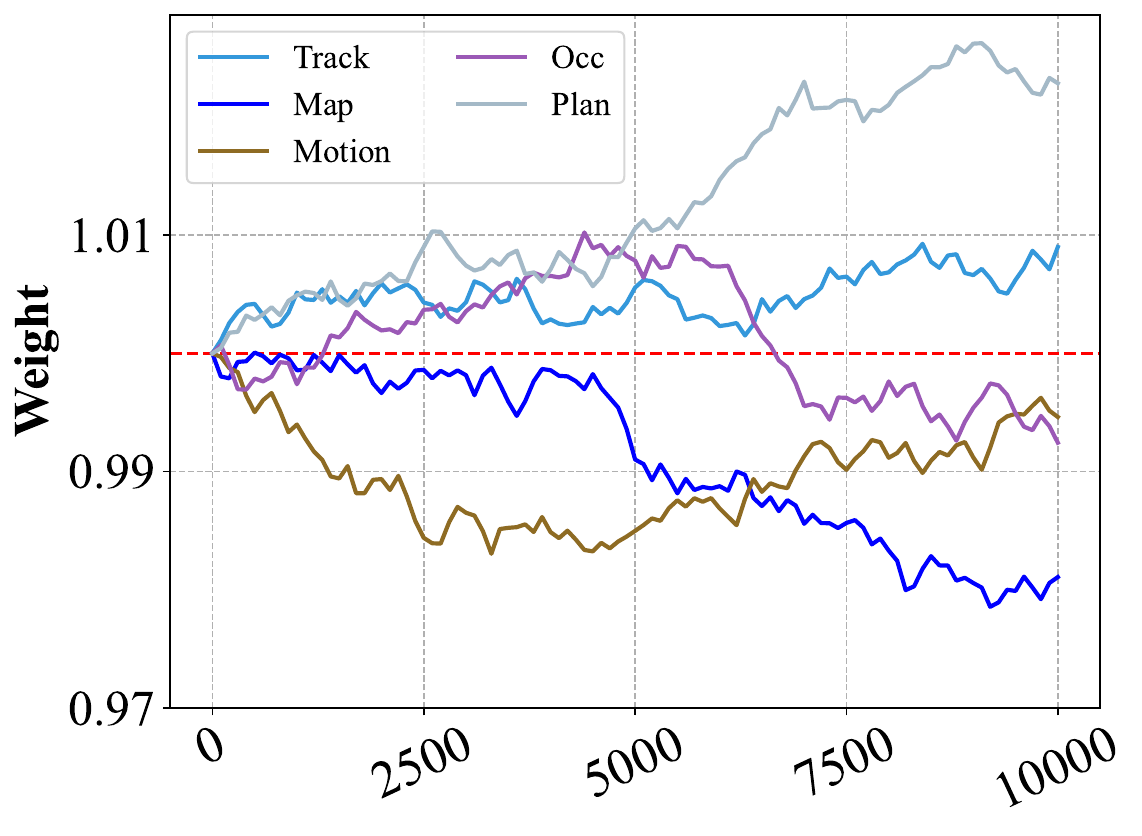}
        \caption{Weight Change}
        \label{subfig:weight-change}
    \end{subfigure}
    \hfill
    \begin{subfigure}[b]{0.32\textwidth}
        \centering
        \includegraphics[width=\textwidth]{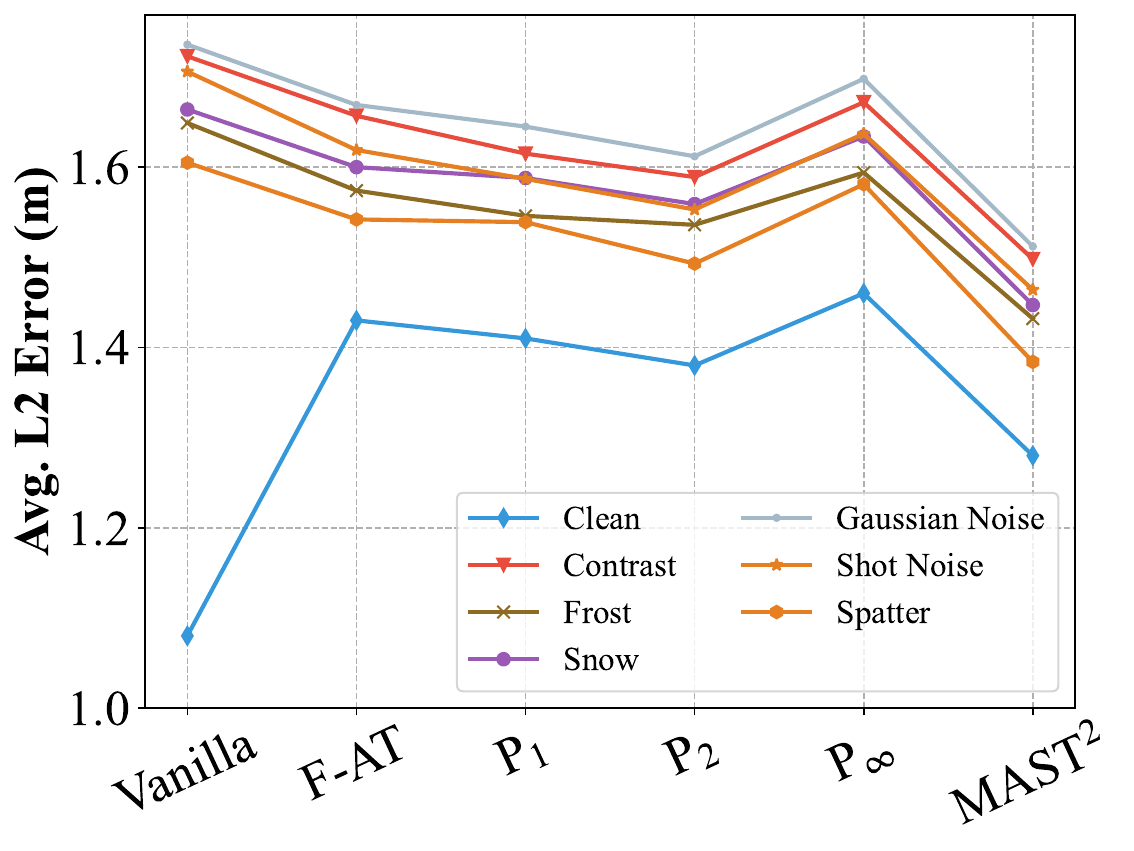}
        \caption{Natural Corruption}
        \label{subfig:natural-corruption}
    \end{subfigure}

\caption{Results of the Discussion Section}
\end{figure*}

To better understand the contributions of different modules, we selectively freeze specific module parameters during training. By keeping these modules' weights fixed, we can assess their impact on the model's overall performance. Using UniAD as the baseline, we freeze five modules and apply $\ell_{\infty}$ perturbation constraints for each (\ie, Track, Map, Motion, Occ, Plan) with values set to 0.8, 0.1, 0.1, 0.1, and 0.1, respectively. The attack steps are set to 5, and fine-tuning is performed over 3 epochs. The results are shown in \Fref{subfig:module-freeze}.

The experimental results reveal significant differences in the impact of freezing different modules. For modules close to the input layer, such as tracks, freezing has little effect and may even improve the model's overall performance. In contrast, freezing modules closer to the output layer, such as Plan, significantly reduces model performance.

\subsection{Dynamic Weight Analysis}
We tracked the weight changes of five modules during \tool training process of UniAD, as shown in \Fref{subfig:weight-change}, where the horizontal axis represents the number of batches. Initially, all module weights are set to 1.0, but they adjust over time. The weights of the Plan and Track modules increase, reaching 1.023 and 1.009, respectively, while the Map module drops to 0.981. This indicates that the Map module's performance declines more rapidly than the others, prompting the Dynamic Weight Accumulation Adaptation to lower its weight for balance. Additionally, module weights fluctuate, with significant changes observed between batches 6500, reflecting the varying contributions of different modules throughout training. Specifically, the Occ module increases rapidly at first, then decreases in later process.

\subsection{Natural Corruption Analysis}
Natural corruption refers to noise commonly encountered in the real world, which differs significantly from adversarial noise. We conducted additional experiments focusing on natural corruption to assess our model's robustness against such disturbances. Following the methodology in \cite{hendrycks2019benchmarking_imagenet-c}, we evaluated our model under various natural corruption scenarios. Specifically, we selected six types of natural corruptions—Contrast, Frost, Snow, Gaussian Noise, Shot Noise, and Spatter—which represent the four distinct corruption categories defined in \cite{hendrycks2019benchmarking_imagenet-c}. These evaluations were performed using the model trained in our main experiment. The results, shown in \Fref{subfig:natural-corruption}, demonstrate that our \tool method consistently performs well against natural corruptions, outperforming both the vanilla model and other adversarial training methods (average drop in Avg. L2 Error of 0.15). This underscores the effectiveness of our approach in defending against natural corruption.

\begin{figure}[ht]
    \centering
    \includegraphics[width=0.99\linewidth]{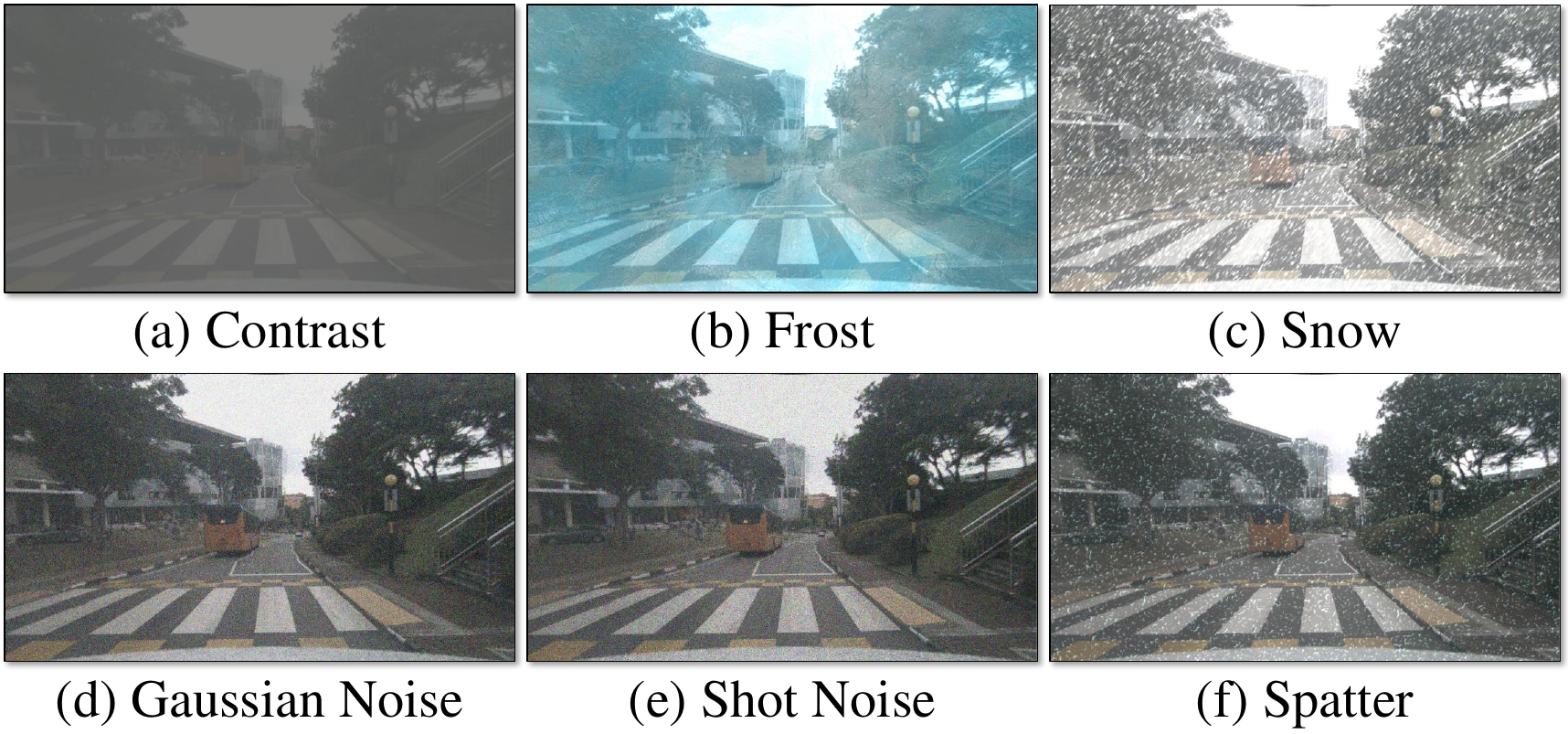}
    \caption{Visual examples for evaluating the robustness of our \tool to natural corruption.}
    \label{fig:d2}
\end{figure}

\subsection{Visualization}

\begin{figure*}[ht]
\centering
	\begin{subfigure}{0.98\linewidth}
 \captionsetup{skip=2pt}
		\centering
		\includegraphics[width=0.98\linewidth]{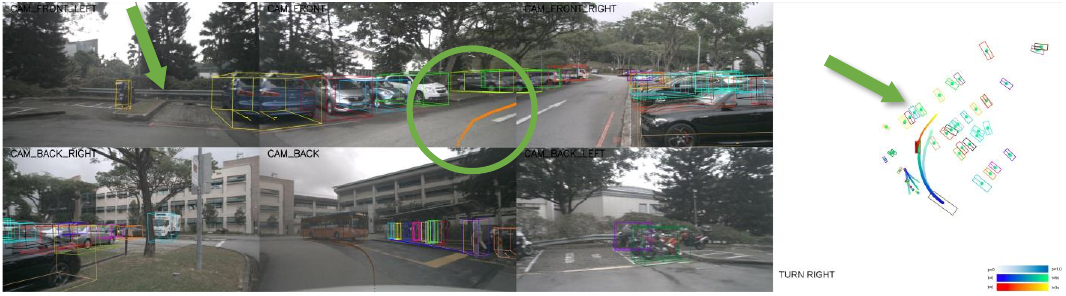}
		\caption{Vanilla UniAD under clean conditions. The model is capable of completing excellent driving tasks.}
		\label{subfig:vis1-ori}%文中引用该图片代号
	\end{subfigure}
 
	\begin{subfigure}{0.98\linewidth}
  \captionsetup{skip=2pt}
		\centering
		\includegraphics[width=0.98\linewidth]{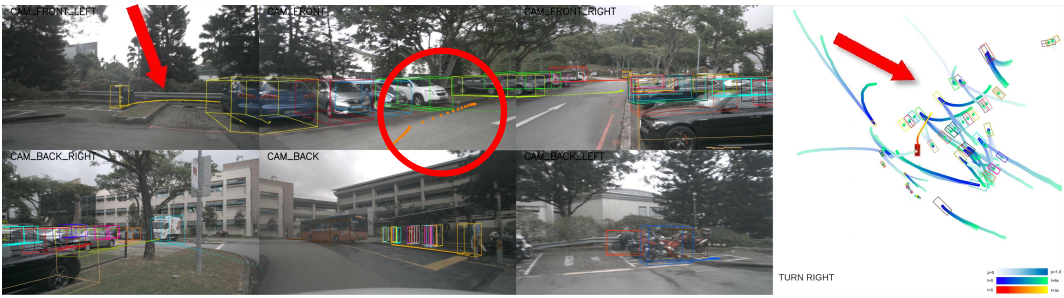}
		\caption{Vanilla UniAD after the PGD-$\ell_\infty$ attack. Most vehicles parked at a standstill on the roadside are predicted to be in motion, as shown in the blue trajectory in the top view on the right. The planned route will collide with vehicles on the left side, as shown in the red circle.}
		\label{subfig:vis1-attack}%文中引用该图片代号
	\end{subfigure}
        \begin{subfigure}{0.98\linewidth}
         \captionsetup{skip=2pt}
		\centering
		\includegraphics[width=0.98\linewidth]{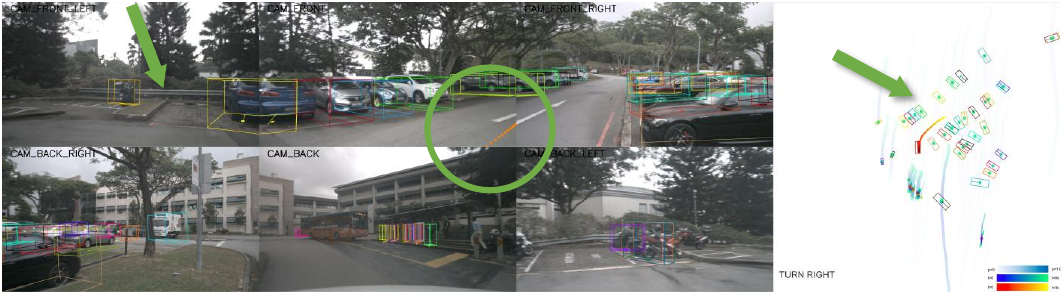}
		\caption{\enhanced UniAD after the PGD-$\ell_\infty$ attack. The vehicles parked on both sides of the road are correctly predicted and the planned driving route is located in the safe center of the road, as shown in the green circle.}
		\label{subfig:vis1-dedfense}%文中引用该图片代号
	\end{subfigure}
\caption{The visual comparison in the same scenario before and after the defense of UniAD. \toolns effectively fixes errors in perception, prediction, and planning, especially in the labeled circles and arrows, where \textcolor{green}{green} represents correct detection, prediction, and planning results, while \textcolor{red}{red} represents incorrect ones.}
\label{fig:vis-comp}
\end{figure*}

We also report the visualization comparison of key driving scenarios. \Fref{fig:vis-comp} illustrates (a) vanilla UniAD under clean conditions, (b) vanilla UniAD after the PGD-$\ell_\infty$ attack, and (c) \enhanced UniAD after the PGD-$\ell_\infty$ attack. In this scenario, UniAD makes critical errors in predicting the trajectories of surrounding vehicles, incorrectly identifying many stationary vehicles as being in motion (as indicated by the red arrow in the BEV view on the right side of \Fref{subfig:vis1-attack}). This leads to the ego vehicle mistakenly avoiding future vehicles that don't actually exist. Consequently, UniAD's final trajectory collides with a stationary vehicle on the left side (as shown by the red circle in \Fref{subfig:vis1-attack}). However, after applying \toolns, the model accurately predicts the future motion states of surrounding vehicles, closely resembling the scenario without an attack. The resulting decision trajectory remains centered on the road (as indicated by the blue circle in \Fref{subfig:vis1-dedfense}).

\section{Related Works}

Adversarial attacks present significant challenges in AD, particularly when targeting advanced end-to-end models. Recent studies aim at evaluating and enhancing the robustness of AD systems against such attacks. For instance, Zhang \etal \cite{zhang2022adversarial} demonstrate how minor adversarial attacks can severely impact trajectory prediction. Similarly, Cao \etal \cite{wu2023adversarial} highlight that even robust end-to-end models are susceptible to adversarial perturbations, which can significantly compromise safety. Furthermore, Zhang \etal \cite{chahe2023dynamic} investigate dynamic adversarial attacks that adapt to environmental changes, emphasizing the challenges of defending against these threats in real-time scenarios.

Adversarial training is currently one of the most effective methods for enhancing the robustness of neural networks. As autonomous driving technology advances, the role of adversarial training in fortifying model resilience within this field becomes increasingly critical. Li \etal \cite{li2023advmono3d} introduced a depth-aware adversarial training method (AdvMono3D) for monocular 3D object detection, which improves model robustness in complex scenarios by incorporating multi-scale adversarial perturbations. Similarly, Li \etal \cite{li2023among} employed an adversarial training strategy that strengthens collaborative perception systems against adversarial attacks through a consensus mechanism, thereby enhancing the safety of autonomous driving. Zhang \etal \cite{zhang2024comprehensive} conducted a comprehensive analysis of the vulnerability of LiDAR 3D object detectors to adversarial attacks in autonomous driving, integrating adversarial training methods to bolster model robustness. Wang \etal \cite{wang2022adversarial} enhanced the resilience of multi-sensor fusion systems through adversarial training, enabling these systems to more effectively resist various adversarial attacks, thereby improving the overall robustness and safety of autonomous driving systems.

\textbf{Differences.} As discussed above, current adversarial training research for autonomous driving models primarily focuses on single tasks or modules, such as 3D object detection and other sub-tasks. This paper focuses on adversarial training in end-to-end autonomous driving models and proposes Module-wise Adaptive Adversarial Training (\toolns), the first adversarial training method specifically designed for end-to-end autonomous driving models, addressing a gap in existing research. Additionally, we innovatively introduce modular noise injection and a dynamic weight accumulation adaptive mechanism, addressing the challenges of inconsistent objectives across modules and strong inter-module dependencies in end-to-end models, significantly enhancing the model's robustness in various adversarial attack scenarios.

\section{conclusion and outlook}
End-to-end autonomous driving (AD) models, which integrate perception, prediction, and planning into a unified framework, offer significant advantages in simplifying decision-making processes. However, their tightly coupled nature also makes them particularly susceptible to adversarial perturbations, and the lack of comprehensive adversarial training methods leaves these models vulnerable to attacks. Existing defenses typically focus on individual tasks within the AD pipeline and are often limited to specific types of perturbations, failing to address the complexity and interconnectedness of end-to-end AD systems.

In this paper, we introduced Module-wise Adaptive Adversarial Training (\toolns), a novel approach specifically designed to enhance the robustness of end-to-end AD models against a wide range of adversarial attacks. \tool addresses the unique challenges of these models by incorporating module-wise noise injection and dynamic weight accumulation adaptation, ensuring balanced and effective training across all stages of the AD pipeline.

We demonstrated the effectiveness of \tool through extensive experiments on the nuScenes dataset, where it significantly outperformed existing adversarial training methods across multiple tasks. Furthermore, closed-loop evaluations in the CARLA simulator confirmed that \tool improves the robustness of end-to-end AD models in closed-loop evaluation.

\textbf{Limitations.} Despite the promising results, several areas remain to be explored: \ding{182} evaluating \tool on real-world vehicles to assess its effectiveness in practical autonomous driving scenarios; \ding{183} developing more advanced adversarial training strategies that can further enhance robustness against a wider range of attacks; and \ding{184} reducing the complexity of the model and accelerating the training process to make \tool more feasible for deployment in real-time systems.

% {\appendix[]
% \input{appendices/appendix1}
% }

\section*{Acknowledgments}
This work was supported by the National Natural Science Foundation of China (Grant. 62206009), State Key Laboratory of Complex \& Critical Software Environment (CCSE), Aeronautical Science Fund (Grant. 20230017051001), and Outstanding Research Project of Shen Yuan Honors College, BUAA (Grant. 230123206).

% {\appendices
% \input{appendices/appendix1}
% }

% {\appendices
% \section*{Proof of the First Zonklar Equation}
% Appendix one text goes here.
% You can choose not to have a title for an appendix if you want by leaving the argument blank
% \section*{Proof of the Second Zonklar Equation}
% Appendix two text goes here.}

% \section{References Section}
% You can use a bibliography generated by BibTeX as a .bbl file.
%  BibTeX documentation can be easily obtained at:
%  http://mirror.ctan.org/biblio/bibtex/contrib/doc/
%  The IEEEtran BibTeX style support page is:
%  http://www.michaelshell.org/tex/ieeetran/bibtex/
 
 % argument is your BibTeX string definitions and bibliography database(s)
%\bibliography{IEEEabrv,../bib/paper}

\bibliographystyle{IEEEtran}
\bibliography{sample-base}

% \newpage

% \section{Biography Section}

% \begin{IEEEbiography}[{\includegraphics[width=1in,height=1.25in,clip,keepaspectratio]{photos/logo1.png}}]{Haha}
% is him.
% \end{IEEEbiography}

% \vfill

% \begin{IEEEbiography}
% [{\includegraphics[width=1in,height=1.25in,clip,keepaspectratio]{photos/logo2.png}}]{Michael Shell}
% is her.
% \end{IEEEbiography}

\end{document}